%% file: main.tex
\documentclass{article}

\input{sections/preamble}
\usepackage[final]{neurips_2024}

\title{
    AdapTable: Test-Time Adaptation for Tabular Data \\ via Shift-Aware Uncertainty Calibrator \\ and Label Distribution Handler
}
\author{%
    Changhun Kim$^{1,2*}$ \hquad Taewon Kim$^{1*}$ \hquad Seungyeon Woo$^{1,3}$ \hquad June Yong Yang$^{1}$ \hquad Eunho Yang$^{1,2}$ \\
    $^1$Korea Advanced Institute of Science and Technology (KAIST) \hquad $^2$AITRICS \hquad $^3$Skelter Labs \\
    \texttt{\{changhun.kim, maxkim139, oox1987, laoconeth, eunhoy\}@kaist.ac.kr} \\
}

\begin{document}
\maketitle
\def\thefootnote{*}\footnotetext{Equal contribution.}

\input{sections/0_abstract}
\input{sections/1_introduction}
\input{sections/2_analysis}
\input{sections/3_method}
\input{sections/4_experiments}
\input{sections/6_conclusion}
\input{sections/7_acknowledgments}
\bibliography{sections/reference}
\input{sections/8_appendix}

\end{document}

%% file: sections/preamble.tex
\usepackage[utf8]{inputenc}
\usepackage[T1]{fontenc}
\usepackage{hyperref}
\usepackage{url}
\usepackage{booktabs}
\usepackage{amsfonts}
\usepackage{nicefrac}
\usepackage[verbose]{microtype}
\usepackage{xcolor}
\usepackage{times}
\usepackage{helvet}
\usepackage{courier}
\usepackage{graphicx}
\PassOptionsToPackage{numbers,compress}{natbib}
\usepackage{natbib}
\usepackage[labelfont=bf]{caption}
\usepackage{algorithm}
\usepackage[english]{babel}

\usepackage{kotex}
\usepackage{newfloat}
\usepackage{listings}
\usepackage{arydshln}
\usepackage{multirow}
\usepackage{tabularx}
\usepackage{makecell}
\usepackage{amsmath}
\usepackage{kotex}
\usepackage{lipsum}
\usepackage{pifont}
\usepackage{relsize}
\usepackage{tikz}
\usepackage{eqparbox}
\usepackage{verbatim}
\usepackage{tabularray}
\usepackage{algpseudocode}
\usepackage{color}
\usepackage{colortbl}
\usepackage{adjustbox}
\usepackage{appendix}
\usepackage{bm}
\usepackage{amssymb}
\usepackage{siunitx}
\usepackage{amsthm}
\usepackage{mathtools}

\DeclareFontFamily{U}{stix2bb}{}
\DeclareFontShape{U}{stix2bb}{m}{n} {<-> stix2-mathbb}{}

\DeclareMathAlphabet{\mathbbold}{U}{bbold}{m}{n}

\def\eqref#1{equation~\ref{#1}}
\def\1{\bm{1}}

\def\vh{{\bm{h}}}

\def\mA{{\bm{A}}}

\DeclareMathAlphabet{\mathsfit}{\encodingdefault}{\sfdefault}{m}{sl}
\SetMathAlphabet{\mathsfit}{bold}{\encodingdefault}{\sfdefault}{bx}{n}

\DeclareMathOperator*{\argmax}{arg\,max}

\newcommand{\hquad}{\hspace{0.5em}}

\newcommand{\ie}{\textit{i}.\textit{e}.}

\newcommand*\colourcheck[1]{
  \expandafter\newcommand\csname #1check\endcsname{\textcolor{#1}{\ding{51}}}%
}
\newcommand*\colourx[1]{
  \expandafter\newcommand\csname #1x\endcsname{\textcolor{#1}{\ding{55}}}%
}
\newcommand{\BF}[1]{\textbf{#1}}

\definecolor{custompink}{rgb}{0.99, 0.03, 0.51}
\colourx{custompink}
\definecolor{brightgray}{gray}{0.9}
\definecolor{lightgray}{gray}{0.8}

\newtheorem{thm}{Theorem}[section]
\newtheorem{defn}[thm]{Definition}

\algrenewcommand\algorithmiccomment[1]{\hfill\textcolor{gray}{$\triangleright$ #1}}

%% file: sections/0_abstract.tex
\begin{abstract}

In real-world scenarios, tabular data often suffer from distribution shifts that threaten the performance of machine learning models. Despite its prevalence and importance, handling distribution shifts in the tabular domain remains underexplored due to the inherent challenges within the tabular data itself.
In this sense, test-time adaptation (TTA) offers a promising solution by adapting models to target data without accessing source data, crucial for privacy-sensitive tabular domains. However, existing TTA methods either 1) overlook the nature of tabular distribution shifts, often involving label distribution shifts, or 2) impose architectural constraints on the model, leading to a lack of applicability.
To this end, we propose AdapTable, a novel TTA framework for tabular data. AdapTable operates in two stages: 1) calibrating model predictions using a shift-aware uncertainty calibrator, and 2) adjusting these predictions to match the target label distribution with a label distribution handler.
We validate the effectiveness of AdapTable through theoretical analysis and extensive experiments on various distribution shift scenarios. Our results demonstrate AdapTable’s ability to handle various real-world distribution shifts, achieving up to a 16\% improvement on the HELOC dataset. Our code is available at \url{https://github.com/drumpt/AdapTable}.

\end{abstract}

%% file: sections/1_introduction.tex
\section{Introduction}

Tabular data is one of the most abundant forms across various industries, including healthcare~\citep{johnson2016mimic}, finance~\citep{FICO}, manufacturing~\citep{manufacturing}, and public administration~\cite{tableshift}. However, tabular learning models often face challenges in real-world applications due to distribution shifts, which severely degrade their integrity and reliability. In this regard, test-time adaptation (TTA)~\citep{pl,liu2021ttt++,wang2020tent,niu2022efficient,niu2023towards,lame} offers a promising solution to address this issue by adapting models under unknown distribution shifts using only unlabeled test data without access to training data.

Despite its potential, direct application of TTA without the consideration of characteristics of tabular data, results in limited performance gain or model collapse. We identify two primary reasons for this. First, representation learning in the tabular domain is often hindered by the entanglement of covariate shifts and concept shifts~\citep{whyshift}. Consequently, TTA methods leveraging unsupervised objectives, which rely on cluster assumption often fail or lead to model collapse. Second, these approaches often do not take label distribution shifts into account, a key factor in the performance decline within the tabular domain. This issue is further aggravated by the tendency for predictions in the target domain to be biased towards the source label.

To address these issues, we propose AdapTable, a novel TTA approach tailored for tabular data. AdapTable consists of two main components: 1) a shift-aware uncertainty calibrator and 2) a label distribution handler. Our shift-aware uncertainty calibrator utilizes graph neural networks to assign per-sample temperature for each model prediction. By treating each column as a node, it captures not only individual feature shifts but also complex patterns across features.
Our label distribution handler then adjusts the calibrated model probabilities by estimating the label distribution of the current target batch. This process aligns predictions with the target label distribution, addressing biases towards the source distribution.
AdapTable requires no parameter updates, making it model-agnostic and thus compatible with both deep learning models and gradient-boosted decision trees, offering high versatility for tabular data. % Additionally, it requires no gradient steps during adaptation, ensuring high efficiency during test time.

We evaluate AdapTable under various distribution shifts and demonstrate AdapTable consistently outperforms baselines, achieving up to 16\% gains on the HELOC dataset. Furthermore, we provide theoretical insights into AdapTable’s performance, supported by extensive ablation studies.
% Furthermore, we also provide theoretical insights into AdapTable’s performance improvements, and supported by ablation studies. % that confirm the individual contributions of the shift-aware uncertainty calibrator and the label distribution handler.
We hereby summarize our contributions:

\begin{itemize}
    \item We analyze the challenges of tabular distribution shifts to reveal why existing TTA methods fail, highlighting the entanglement of covariate, concept shifts, and label distribution shifts as key factors in performance degradation. % This unvails why existing TTA methods fail. % and underscores the need for tabular-specific approaches to handle label distribution shifts.
    
    \item Building on these analyses, we introduce AdapTable, a first model-agnostic TTA method specifically designed for tabular data. AdapTable addresses label distribution shifts by estimating and adjusting the label distribution of the current test batch, while also calibrating model predictions with a shift-aware uncertainty calibrator.
    
    \item Our extensive experiments demonstrate that AdapTable exhibits robust adaptation performance across various model architectures and under diverse natural distribution shifts and common corruptions, further supported by extensive ablation studies.

\end{itemize}

%% file: sections/2_analysis.tex
\section{Analysis of Tabular Distribution Shifts} \label{sec:analysis}

In this section, we examine why prior TTA methods struggle with distribution shifts in the tabular domain. First, we note that deep learning models' latent representations do not follow label-based cluster assumptions due to the entanglement of covariate and concept shifts, causing TTA methods relying on these assumptions~\cite{wang2020tent,pl,niu2022efficient} to falter in tabular data. 
Second, we identify label distribution shift as a key driver of performance degradation under distribution shifts, as discussed further in Section~\ref{subsec:em_failure} and Section~\ref{subsec:label_distribution_shift}.

% In this section, we analyze the limitations of prior TTA works in handling distribution shifts within the tabular domain. First, we highlight that the latent representation of deep learning models does not follow the cluster assumption by label due to the entanglement of covariate and concept shifts. Thus, TTA methods that rely on cluster assumptions~\cite{wang2020tent,pl,niu2022efficient} fail to bring improvements when naively applied to tabular data.
% Second, we find label shift to be a fundamental reason for performance degradation in the face of distribution shifts. We further elaborate them below in Section~\ref{subsec:em_failure} and Section~\ref{subsec:label_distribution_shift}, respectively. 

\subsection{Indistinguishable Representations} \label{subsec:em_failure}

\input{figures/latent_reliability}

We first reveal that deep tabular models fail to learn distinguishable embeddings. 
In Figures~\ref{fig:latent_reliability} (a) and (b), we visualize the embedding spaces of models trained on two datasets: HELOC~\cite{tableshift}, a pure tabular dataset, and Optdigits~\cite{openml_benchmark}, a linearized image dataset. Notably, the deep learning models' representations adhere to the cluster assumption by labels only in the image data, not in the tabular data.
% In Figure~\ref{fig:latent_reliability} (a) and (b), we visualize the embeddings space of models trained on two representative datasets---HELOC~\cite{tableshift}, a pure tabular dataset where each columns represent distinct meanings, and Optdigits~\cite{openml_benchmark}, a linearized image dataset often used in tabular learning. Interestingly, we find that the latent space is ambiguous \emph{only in tabular data}. % Furthermore, as shown in Figure~\ref{fig:latent_reliability} (c) and (d), unlike in the vision domain where trained models exhibit strong overconfidence, tabular models, depending on the dataset, exhibit both overconfidence and underconfidence.

We attribute this unique behavior of deep tabular models to the high-frequency nature of tabular data. In the tabular domain, weak causality from inputs $X$ to outputs $Y$ due to latent confounders $Z$ often leads to vastly different labels for similar inputs~\cite{tree,whyshift}. For instance, cardiovascular disease risk predictions based on cholesterol, blood pressure, age, and smoking history are influenced by gender as a latent confounder, resulting in different risk levels for men and women despite identical inputs~\cite{mosca2011sex,defilippis2021time}. This leads to high-frequency functions that are difficult for deep neural networks, which are biased toward low-frequency functions, to accurately model~\cite{beyazit2024inductive}.

Consequently, prior TTA methods, which rely on cluster assumptions and primarily target input covariate shifts, show limited performance gains. Figure~\ref{fig:various_architectures} demonstrates that these methods fail to improve beyond the vanilla performance of the source model due to the lack of a cluster assumption.

\subsection{Importance of Label Distribution Shifts} \label{subsec:label_distribution_shift}

\input{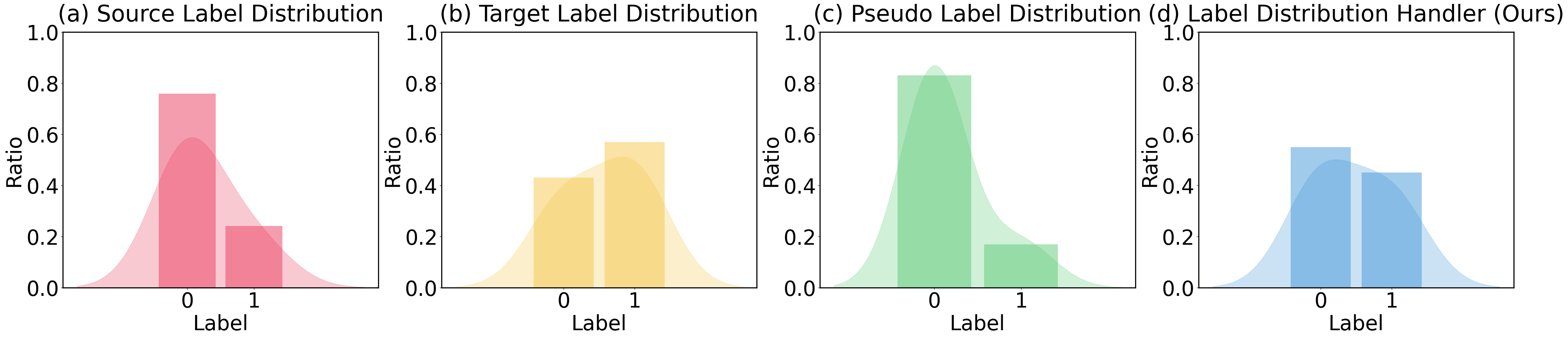}

Second, we find that label distribution shift is a primary cause of performance degradation, and accurate estimation of target label distribution can lead to significant performance gains. A recent benchmark study, TableShift~\cite{tableshift} have emphasized that label distribution shift is a primary cause of performance degradation in tabular data. Specifically, They investigated the relationship between three key shift factors---input covariate shift ($X$-shift), concept shift ($Y|X$-shift), and label distribution shift ($Y$-shift)---and model performance, and discovered that label distribution shifts are strongly correlated with performance degradation. Our analysis in Section~\ref{sec:further_analysis} further reveals that these shifts are highly prevalent in tabular data. This underscores the need for a test-time adaptation method that addresses label distribution shifts by estimating the target label distribution and adjusting predictions accordingly.

Moreover, we visualize model predictions in Figure~\ref{fig:label_distribution_shift} and observe that, similar to other domains~\cite{wu2021online,hwang2022combating,park2023pc}, the marginal distribution of output labels is biased toward the source label distribution. Given that tabular models are often poorly calibrated (Figure~\ref{fig:latent_reliability}), we conduct an experiment using a perfectly calibrated model, which yields high confidence for correct samples and low confidence for incorrect ones. As shown in Table~\ref{table:observation_oracle}, our label distribution adaptation method significantly improves under these conditions. This underscores the need for an uncertainty calibrator specific to tabular data.

\input{tables/observation_oracle}

%% file: figures/latent_reliability.tex
\begin{figure}[!t]
\centering
\includegraphics[width=.65\linewidth]{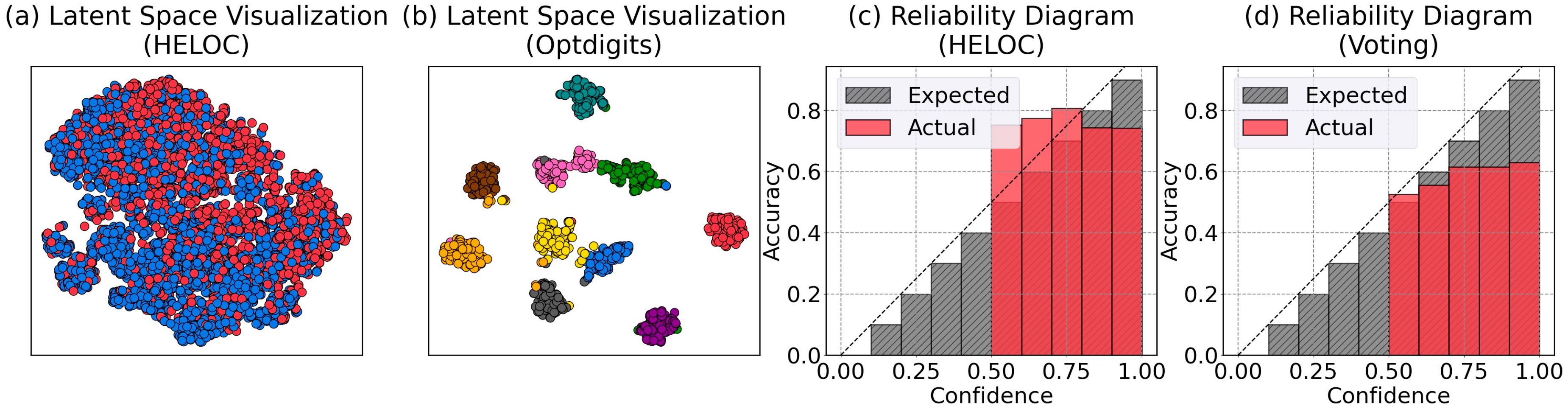}
\caption{
    Latent space visualization with t-SNE comparing (a) tabular data~\cite{tableshift} and (b) image data~\cite{openml_benchmark}. Reliability diagrams of (c) underconfident and (d) overconfident scenarios are shown. All experiments are conducted using an MLP architecture.
}
\label{fig:latent_reliability}
\vspace{-.3in}
\end{figure}

%% file: figures/label_distribution_shift.tex
\begin{figure}[!t]
\centering
\includegraphics[width=.65\linewidth]{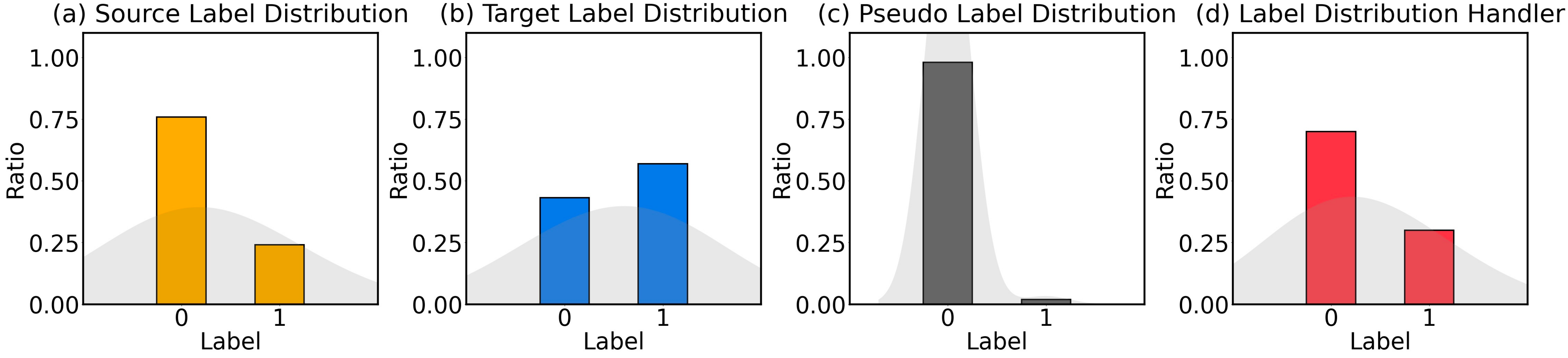}
\caption{
    Label distribution of (a) source domain, (b) target domain, (c) estimated label distribution using pseudo labels, and (d) corrected label distribution of AdapTable are shown using MLP on HELOC dataset.
}
\label{fig:label_distribution_shift}
\vspace{-.3in}
\end{figure}

%% file: tables/observation_oracle.tex
\begin{table}[!ht]
\vspace{-.2in}
\centering
\caption{
    Key findings demonstrate that uncertainty calibration enhances the performance of the label distribution handler. % The best results are highlighted in \textbf{bold}, and the second-best results are \underline{underlined}.
}
\label{table:observation_oracle}
\begin{adjustbox}{width=.3\linewidth}
\setlength{\tabcolsep}{2pt}
\begin{tabular}{lcc}
    \toprule
    
    Method & {HELOC} & {Voting} \\ \midrule
    
    Source & 47.6 & 79.3 \\
    
    \rowcolor{brightgray} AdapTable & \underline{63.7} & \underline{79.6} \\
    
    \rowcolor{lightgray} AdapTable (Oracle) & \textbf{90.1} & \textbf{84.7} \\    
    
    \bottomrule
\end{tabular}
\end{adjustbox}
\vspace{-.2in}
\end{table}

%% file: sections/3_method.tex
\section{AdapTable} \label{sec:method}

This section introduces AdapTable, the first model-agnostic test-time adaptation strategy for tabular data. AdapTable uses per-sample temperature scaling to correct overconfident yet incorrect predictions; by treating each column as a graph node, it employs a shift-aware uncertainty calibrator with graph neural networks to capture both individual and complex feature shifts~(Section~\ref{subsec:calibrator}). It also estimates the average label distribution of the current test batch and adjusts the model’s output predictions accordingly~(Section~\ref{subsec:label_distribution_handler}).
We also provide a theoretical justification for how our label distribution estimation reduces the error bound in Section~\ref{subsec:thm}. The overall framework of AdapTable is depicted in Figure~\ref{fig:concept_figure}.

\subsection{Test-Time Adaptation Setup for Tabular Data} \label{subsec:setup}

We begin by defining the problem setup for test-time adaptation (TTA) for tabular data. Let $f_{\theta}: \mathbb{R}^{D} \rightarrow \mathbb{R}^{C}$ be a pre-trained classifier on the labeled source tabular domain $\mathcal{D}_s = \{( \mathbf{x}_i^s, y_i^s )\}_{i} \subset X_s \times Y_s$, where each pair consists of a tabular input $\mathbf{x}_i^s \in % X_s \subset
\mathcal{X} = \mathbb{R}^{D}$ and its corresponding output class label $y_i^s \in % Y_s \subset
\mathcal{Y} = \{1, \cdots, C \}$. The classifier takes a row $\mathbf{x}_i \in \mathbb{R}^{D}$ from a table and returns output logit $f_{\theta}(\mathbf{x}_i) \in \mathbb{R}^{C}$. Here, $D$ and $C$ are the number of input features and output classes, respectively. The objective of TTA for tabular data is to adapt $f_{\theta}$ to the unlabeled target tabular domain $\mathcal{D}_t = {\{\mathbf{x}_i^{t} \}}_{i} = X_t$ %\subset X$
during inference, without access to $\mathcal{D}_s$. Unlike most TTA methods that fine-tune model parameters $\theta$ with unsupervised objectives, our approach directly adjusts the output prediction $f_{\theta}(\mathbf{x}_i^{t})$.

\input{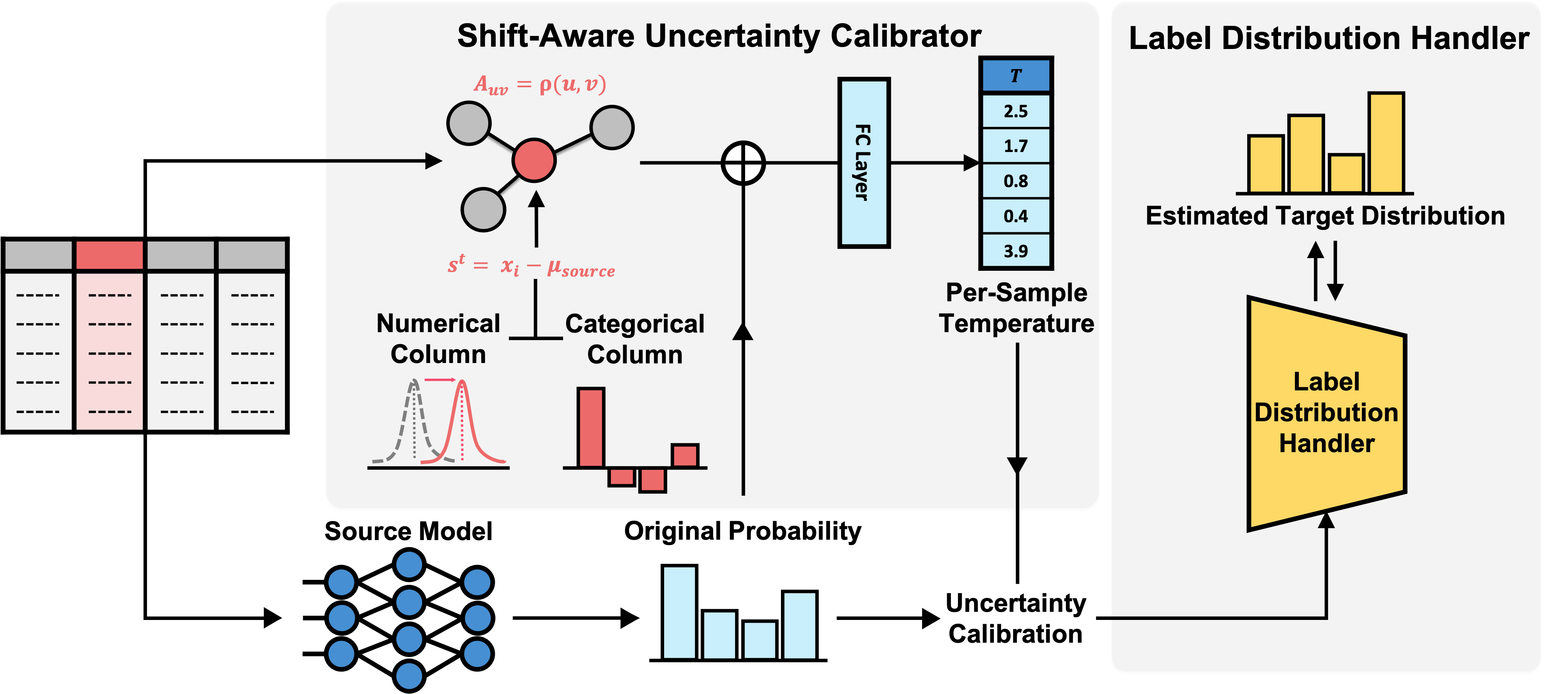}

\subsection{Shift-Aware Uncertainty Calibrator} \label{subsec:calibrator}

This section describes a shift-aware uncertainty calibrator $g_{\phi}: \mathbb{R}^{C} \times \mathbb{R}^{D \times N} \rightarrow \mathbb{R}^{+}$ designed to adjust the poorly calibrated original predictions $p_t(y|\mathbf{x}_i^{t}) = \text{softmax}(f_{\theta}(\mathbf{x}_i^{t}))$, where $\text{softmax}(z)_i = \exp(z_i) / \sum_{i'} \exp(z_{i'})$ normalizes the logits.
Our shift-aware uncertainty calibrator lowers the confidence of overconfident yet incorrect predictions, thereby 1) facilitating better alignment of these predictions with the estimated target label distribution, and 2) mitigating their impact on the inaccurate estimation of the target label distribution.

Conventional post-hoc calibration methods~\cite{platt,isotonic} typically take solely the original model prediction $f_{\theta}(\mathbf{x}_i^{t})$ as input and return the corresponding temperature $T_i$ without taking input variations into account. We argue that this can be suboptimal as it fails to account for the uncertainty arising from variations in the input itself. Instead, our $g_{\phi}$ not only considers $f_{\theta}(\mathbf{x}_i^{t})$ but also incorporates $\mathbf{x}_i^{t}$ with the shift trend $\mathbf{s}^t$ of the current batch as additional inputs. Capturing the common shift patterns within the current batch enables a more accurate reflection of the uncertainty caused by the overall shift patterns within the current batch.

In detail, the shift trend $\mathbf{s}^t = (\mathbf{s}_u^t)_{u=1}^{D} \in \mathbb{R}^{D \times N}$ is defined for a specific column index $u$ as follows:
\begin{equation} \label{eqn:shift_trend}
    \mathbf{s}_{u}^{t} = \big( \mathbf{x}^{t}_{iu} - \frac{1}{|\mathcal{D}_s|} \sum_{i'=1}^{|\mathcal{D}_s|} \mathbf{x}_{i'u}^s \big)_{i=1}^{N} \in \mathbb{R}^{N}.
\end{equation}
Here, $\mathbf{s}_u^t$ represents the difference between the values of the $u$-th column within the current test batch and the average values of the corresponding column in the source data. Using $\mathbf{s}_u^t$ for each column $u$, we define a shift trend graph where each node $u$ represents a column, and each edge captures the relationship between different columns; the node feature for each node $u$ is defined as $\mathbf{s}_u^t$, and the adjacency matrix is represented by an $D \times D$ all-ones matrix.

A graph neural network (GNN) is then applied to the graph formed above, enabling the exchange of shift trends between different columns through message passing. This process generates a column-wise contextualized representation, which is then averaged to produce an overall feature representation that encompasses all columns. Finally, the averaged node representation is concatenated with the initial prediction $f_{\theta}(\mathbf{x}_i^t)$ to yield the final output temperature $T_i$. This GNN-based uncertainty calibration not only captures shifts in individual columns but also sensitively detects correlation shifts occurring simultaneously across different columns, which are common in the tabular domain.
A more detailed explanation of the architecture and training of the shift-aware uncertainty calibrator can be found in Section~\ref{sec:adaptable_detail}.

\subsection{Label Distribution Handler} \label{subsec:label_distribution_handler}

This section introduces a label distribution handler designed to accurately estimate the target label distribution for the current test batch and adjust the model's output predictions accordingly. This approach is empirically justified by our observation that the marginal distribution of model predictions $p_t(y)$ in the target domain tends to be biased towards the source label distribution $p_s(y)$, as discussed in Section~\ref{subsec:label_distribution_shift} and illustrated in Figure~\ref{fig:label_distribution_shift}.

A straightforward solution to correct this bias is to simply multiply $p_t(y) / p_s(y)$ to align the marginal label distribution~\cite{remixmatch}. Specifically, given $p_t(y | \mathbf{x}_{i}^{t}) = \text{softmax}(f_{\theta}(\mathbf{x}_{i}^{t}))$, the adjusted prediction would be:
\begin{equation} \label{eqn:dist_align}
    \text{norm}(p_t(y | \mathbf{x}_{i}^{t}) p_t(y) / p_s(y))
\end{equation}
where $\text{norm}(z)_i = z_i / \sum_{i'}{z_{i'}}$ normalizes the unnormalized probability. However, we find two major issues: 1) $p_t(y | \mathbf{x}_{i}^{t})$ is often poorly calibrated and 2) overconfident yet incorrect predictions significantly hinder the accurate estimation of the target label distribution $p_t(y)$ (Section~\ref{subsec:label_distribution_shift}).

To tackle these challenges, we propose a simple yet effective estimator $\bar{p}_{i}(y|\mathbf{x}_{i}^{t})$ defined like below:
\begin{equation} \label{eqn:final}
    \bar{p}_{i}(y|\mathbf{x}_{i}^{t}) = \frac{
        \tilde{p}_t(y|\mathbf{x}_{i}^{t}) + \text{norm}\big(\tilde{p}_t(y|\mathbf{x}_{i}^{t}) p_t(y) / p_s(y) \big)
    }{2}.
\end{equation}
The key differences between the original Equation~\ref{eqn:dist_align} and our Equation~\ref{eqn:final} are: 1) we use the calibrated prediction $\tilde{p}_t(y|\mathbf{x}_{i}^{t})$ instead of the original prediction $p_t(y | \mathbf{x}_{i}^{t})$ to enhance uncertainty quantification, and 2) we combine the calibrated estimate $\tilde{p}_t(y|\mathbf{x}_{i}^{t})$ with the distributionally aligned prediction $\text{norm}(\tilde{p}_t(y|\mathbf{x}_{i}^{t}) p_t(y) / p_s(y))$ for more robust estimation.

Given the already-known source label distribution $p_s(y)$, we now explain the step-by-step process for estimating $\tilde{p}_t(y|\mathbf{x}_{i}^{t})$ and $p_t(y)$. $p_t(y | \mathbf{x}_{i}^{t})$ is calibrated into $\tilde{p}_t(y|\mathbf{x}_{i}^{t})$ through a two-stage uncertainty calibration process.
Specifically, for a current test batch $\{\mathbf{x}_{i}^{t} \}_{i=1}^{N}$, we calculate shift trend $\mathbf{s}^{t}$ using Equation~\ref{eqn:shift_trend} and get per-sample temperature $T_i = g_{\phi}(f_{\theta}(\mathbf{x}_i^t), \mathbf{s}^t)$ using shift-aware uncertainty calibrator $g_{\phi}$ to capture overall distribution shifts, as well as correlation and individual column shifts within the current batch. Here, we define the uncertainty $\delta_{i}$ of $f_{\theta}(\mathbf{x}_i^t)$ as a reciprocal of the margin of the calibrated probability distribution $\text{softmax}(f_{\theta}(\mathbf{x}_i^t) / T_i)$.
We then measure the quantiles for each instance $\mathbf{x}_{i}$ using $\delta_{i}$ within the current batch and recalibrate the original probability with $\tilde{T}_i$, resulting in $\tilde{p}_t(y|\mathbf{x}_{i}^{t}) = \text{softmax}(f_{\theta}(\mathbf{x}_i^t) / \tilde{T}_i)$. This process calibrates predictions by leveraging relative uncertainty within the batch.
Our temperature $\tilde{T}_i$ is defined as:
\begin{equation} \label{eqn:t_tilde_i}
    \tilde{T}_i =
    \begin{cases}
        T & \text{if } \delta_{i} \ge Q\big(\{\delta_{i'}\}_{i'=1}^{N}, q_{\text{high}}\big) \\
        1 / T & \text{if } \delta_{i} \le Q\big(\{\delta_{i'}\}_{i'=1}^{N}, q_{\text{low}}\big) \\
        1 & \text{otherwise},
    \end{cases}
\end{equation}
where $Q(X, q)$ is a quantile function which gives the value corresponding to the lower $q$ quantile in $X$, $T = 1.5 \rho / (\rho - 1 + 10^{-6})$ is a temperature with source class imbalance ratio $\rho = \max_{j} p_{s}(y)_{j} / \min_{j} p_{s}(y)_{j}$, and $q_{\text{low}}$ and $q_{\text{high}}$ represent the low and high uncertainty quantiles, respectively. This two-stage uncertainty calibration comprehensively evaluates the current batch and estimates relative uncertainty using $\mathbf{s}^{t}$, $g_{\phi}$, and $\tilde{T}_i$.

Meanwhile, the target label distribution $p_t(y)$ is estimated as follows:
\begin{equation}
    p_t(y) = (1 - \alpha) \cdot \frac{1}{N} {\sum_{i=1}^{N} p^{\text{de}}_{t}(y|\mathbf{x}_{i}^{t})} + \alpha \cdot {p^{\text{oe}}_t(y)},
\end{equation}
where $p^{\text{de}}_{t}(y|\mathbf{x}_{i}^{t}) = \text{norm}\big( p_t(y|\mathbf{x}_{i}^{t}) / p_s(y) \big)$ is a debiased target label estimator that departs from $p_s(y)$, and $p_t^{oe}(y)$ is an online target label estimator, initialized as uniform distribution and updated as:
\begin{equation} \label{eqn:online_target_label_estimator}
    p^{\text{oe}}_{t}(y) = (1 - \alpha) \cdot \frac{1}{N} \sum_{i=1}^{N}{\bar{p}_{t}(y|\mathbf{x}_{i}^{t})} + \alpha \cdot p^{\text{oe}}_{t}(y)    
\end{equation}
from the current batch to the next, with a smoothing factor $\alpha$. This online target label estimator leverages label locality between nearby test batches, making it effective for accurately estimating the next batch's target label distribution. A more detailed explanation of AdapTable is provided in Section~\ref{sec:adaptable_detail}.

\subsection{Theoretical Insights} \label{subsec:thm}

\begin{thm} \label{thm:main}
Let $\hat{Y}|X$ and $\hat{Y}_{o}|X$ be defined as follows:
\begin{align}
    \hat{Y}|X &= \{ \argmax_{j \in \mathcal{Y}}{f_{\theta}(\mathbf{x})_{j}}|\mathbf{x} \in X \}, \\
    \hat{Y}_{o}|X &= \{ \argmax_{j \in \mathcal{Y}}{f_{\theta}(\mathbf{x})_{j} + \log p_{t}^{oe}(y)_j}|\mathbf{x} \in X\}.
\end{align}
Given the error $\epsilon(\hat{Y} | X) = \mathbb{P}(\hat{Y} \ne Y | X)$, with true labels $Y$ of inputs $X$, the error gap $| \epsilon(\hat{Y} | X_s) - \epsilon(\hat{Y}_{o}|X_t) |$ is upper bounded by
\begin{equation}
K_1 \Big\| 1 - \frac{p_t^{oe}(y)}{p_t(y)} \Big\|_{1} BSE(\hat{Y}) + K_2 \Delta_{CE}(\hat{Y}),
\end{equation}
where $K_1$ and $K_2$ are constants related to $p_t(y)$, and $p_s(y)$, respectively.
\end{thm}
Theorem~\ref{thm:main} extends Theorem 2.3 in ODS~\cite{ods} to cases where the source label distribution is not uniform. It decomposes the error gap between the original model on the source domain and the adapted model for the target model with $p_{t}^{oe}(y)$ on the target domain into several components. These components include $\| 1 - p_{t}^{oe}(y) / p_{t}(y) \|_1$, which is an error of the estimated target label distribution, $BSE(\hat{Y})$, which reflects the model's performance on the source domain, and $\Delta_{CE}(\hat{Y})$, which measures the generalization of feature representations adapted by the TTA algorithm.
Overall, Theorem~\ref{thm:main} underscores the importance of tracking label distributions and efficiently adapting models to handle label distribution shifts. The detailed explanation and proof of Theorem~\ref{thm:main} can be found in Section~\ref{sec:adaptable_proof}.

%% file: figures/concept_figure.tex
\begin{figure*}[!t]
\centering
\includegraphics[width=.8\textwidth]{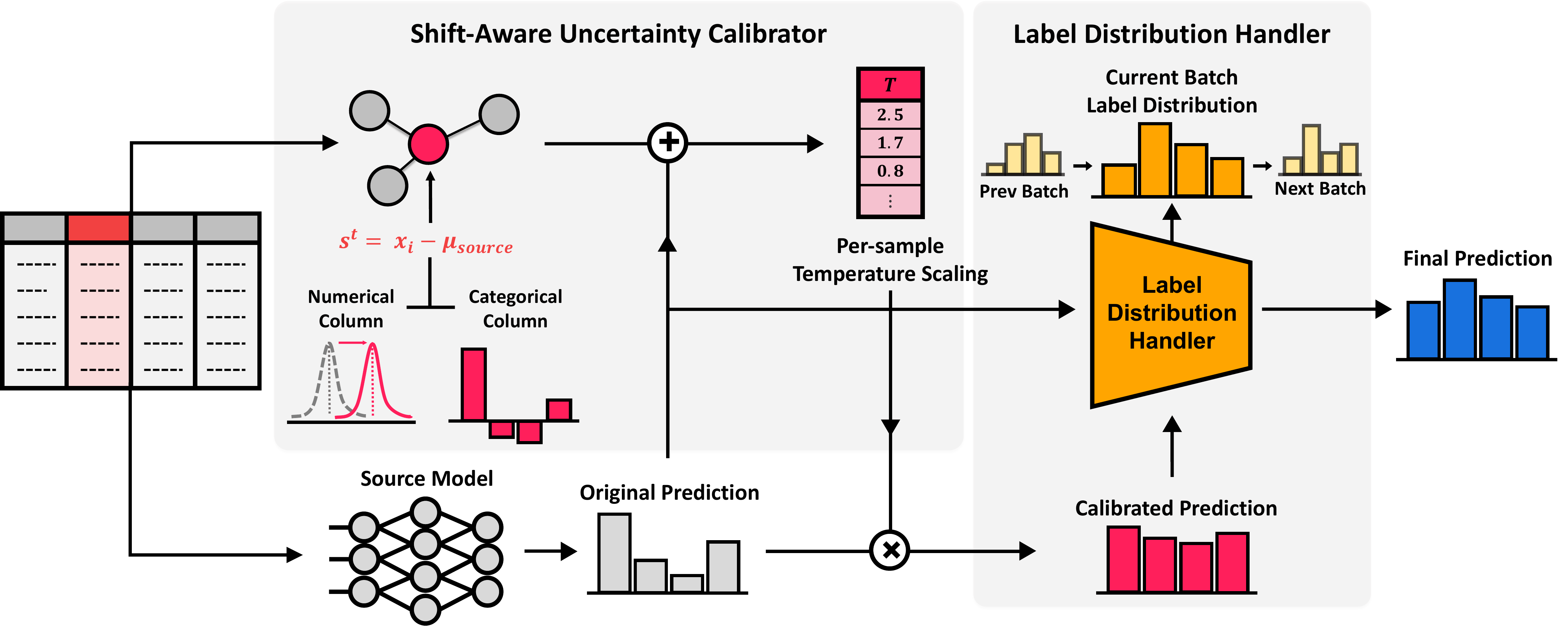}
\caption{
    The overall pipeline of the AdapTable framework.
    AdapTable employs a per-sample temperature scaling to correct overconfident predictions by treating each column as a graph node, enabling a shift-aware uncertainty calibrator with graph neural networks to capture both individual and complex feature shifts (Section~\ref{subsec:calibrator}).
    It also estimates the label distribution of the current test batch and adjusts the model’s predictions accordingly (Section~\ref{subsec:label_distribution_handler}).
}
\label{fig:concept_figure}
\vspace{-.35in}
\end{figure*}

%% file: sections/4_experiments.tex
\section{Experiments} \label{sec:exp}

This section validates AdapTable's effectiveness. We begin with an overview of our experimental setup in Section~\ref{exp:setup} and then address key research questions: Is AdapTable effective across various tabular distribution shifts, including natural shifts and common corruptions across different tabular models? (Section~\ref{exp:main}), Do AdapTable's components contribute to overall performance improvements, and do they function as intended? (Section~\ref{exp:ablation}) Does AdapTable demonstrate strengths in computational efficiency and hyperparameter sensitivity, which are crucial for test time adaptation? (Section~\ref{exp:analysis})

\input{tables/main_natural_distribution_shift}

\subsection{Experimental Setup} \label{exp:setup}

\paragraph{Datasets.}
We evaluate AdapTable on six diverse datasets---HELOC, Voting, Hospital Readmission, ICU Mortality, Childhood Lead, and Diabetes---within the tabular distribution shift benchmark~\citep{tableshift}, covering healthcare, finance, and politics with both numerical and categorical features. Additionally, we verify its robustness against six common corruptions---Gaussian, Uniform, Random Drop, Column Drop, Numerical, and Categorical---to ensure its efficacy beyond label distribution shifts. More details of these shifts are in Section~\ref{sec:dataset_details}.

\paragraph{Model architectures and baselines.}
To verify the proposed method under various tabular model architectures, we mainly use MLP, a widely used tabular learning architecture. Additionally, we validate AdapTable on CatBoost~\citep{catboost} and three other representative deep tabular learning models—AutoInt~\citep{song2019autoint}, ResNet~\citep{fttransformer}, and FT-Transformer~\citep{fttransformer}. We compare AdapTable with six TTA baselines---{PL}~\citep{pl}, {TTT++}~\citep{liu2021ttt++}, {TENT}~\citep{wang2020tent}, {EATA}~\citep{niu2022efficient}, {SAR}~\citep{niu2023towards}, and {LAME}~\citep{lame}. TabLog~\citep{tablog} is excluded due to its architectural constraint on logical neural networks~\citep{riegel2020logical}. We also provide performance references from classical machine learning models: $k$-nearest neighbors ({$k$-NN}), logistic regression ({LogReg}), random forest ({RandomForest}), {XGBoost}~\citep{XGBoost}, and {CatBoost}~\citep{catboost}.

\paragraph{Evaluation metrics and implementation details.}
As shown in Figure~\ref{fig:label_distribution_shift} and Section~\ref{sec:further_analysis}, tabular data often exhibit extreme class imbalance. Since accuracy may not be effective in these cases, we use macro F1 score (F1) and balanced accuracy (bAcc.) as the primary evaluation metrics. For all experiments, we use a fixed batch size of 64, a common setting in TTA baselines~\citep{bnstats,wang2020tent}. The smoothing factor $\alpha$, low uncertainty quantile $q_{\text{low}}$, and high uncertainty quantile $q_{\text{high}}$ are set to 0.1, 0.25, and 0.75, respectively. In all tables, we mark the \textbf{best} and \underline{second-best} results.

\subsection{Main Results} \label{exp:main}

\paragraph{Result on natural distribution shifts.}
Table~\ref{table:main_table} presents results on natural distribution shifts. Existing TTA methods, successful in computer vision, struggle in the tabular domain, often failing to outperform the source model or offering limited performance gains. In contrast, AdapTable achieves state-of-the-art results across all datasets, with dramatic performance improvements of up to 26\% on the HELOC dataset. Since AdapTable does not rely on model parameter tuning, it can be easily applied to classical machine learning models; when integrated with CatBoost, AdapTable consistently improves performance across all datasets, showcasing its versatility, whereas other baselines cannot be similarly integrated as they require model parameter updates.

\paragraph{Result on common corruptions.}
We further evaluate the efficacy of AdapTable across six types of common corruptions in real-world applications by applying them to the test sets of three datasets---HELOC, Voting, and Childhood Lead. As shown in Figure~\ref{fig:main_common_corruption}, prior TTA methods fail considerably, showing only marginal gains over the unadapted source model across all corruption types. It is worth noting that previous TTA methods have demonstrated significant improvements when dealing with common corruptions in vision data, highlighting the difference between corruptions in the tabular domain its counterpart in vision domain. Meanwhile, Adaptable shows substantial improvements across all types of corruptions, showing more than 10\% gains of accuracy on all scenarios, demonstrating its robustness across different types of corruptions.

\input{figures/main_common_corruption}
\input{figures/calibration_efficacy}
\input{tables/ablation_uncertainty_calibrator}

\subsection{Ablation Study} \label{exp:ablation}

\paragraph{Shift-aware uncertainty calibrator.}
We first validate the shift-aware uncertainty calibrator from Section~\ref{subsec:calibrator}. Figures~\ref{fig:calibration_efficacy}(a) and (b) show reliability diagrams before and after calibration, demonstrating that our calibrator significantly reduces both overconfidence and underconfidence. Next, Figure~\ref{fig:calibration_efficacy}(c) assesses shift-awareness by plotting the average temperature against the maximum mean discrepancy (MMD) with training data. As expected, greater shifts lead to higher temperatures, indicating increased uncertainty. The strong correlation between MMD and average temperature confirms the calibrator’s effectiveness under distribution shifts. Finally, Table~\ref{table:ablation_uncertainty_calibrator} compares our calibrator with classical methods like Platt scaling and isotonic regression. While classical methods show inconsistent performance across datasets, our shift-aware calibrator consistently outperforms them, effectively handling domain shifts during calibration.

\input{figures/js_div}

\paragraph{Label distribution handler.}
We next validate the label distribution handler's efficacy by first comparing the Jensen–Shannon (JS) divergence between true and estimated label distributions across online batches in Figure~\ref{fig:js_div}. The results show that our handler significantly improves label distribution estimation accuracy, with low JS divergence across all datasets. We then assess its robustness under severe class imbalance (ratio of 10) and class-wise temporal correlation. As shown in Table~\ref{table:label_distribution_shift}, AdapTable achieves up to 27\% and 19\% performance improvements in the HELOC and Childhood Lead datasets, respectively. More experimental details are in Section~\ref{sec:dataset_details}.

\input{tables/label_distribution_shift}

%% file: tables/main_natural_distribution_shift.tex
\begin{table*}[!t]
\caption{
    The average balanced accuracy (\%) and macro F1 score (\%) with their standard errors for both supervised models and TTA baselines are reported across six datasets including natural distribution shifts within the TableShift~\cite{tableshift} benchmark. The results are averaged over three random repetitions.
}
\label{table:main_table}
\vspace{-.1in}
\centering
\begin{small}
\begin{footnotesize}
\setlength{\columnsep}{1pt}
\begin{adjustbox}{width=.95\linewidth}
\begin{tabular}{lccccccccccccc}
    \toprule
    
    & \multicolumn{2}{c}{{HELOC}} & \multicolumn{2}{c}{{Voting}} & \multicolumn{2}{c}{{Hospital Readmission}} & \multicolumn{2}{c}{{ICU Mortality}} &  \multicolumn{2}{c}{{Childhood Lead}} & \multicolumn{2}{c}{{Diabetes}} \\ 
    
    \cmidrule(l{3pt}r{3pt}){2-3} \cmidrule(l{3pt}r{3pt}){4-5} \cmidrule(l{3pt}r{3pt}){6-7} \cmidrule(l{3pt}r{3pt}){8-9} \cmidrule(l{3pt}r{3pt}){10-11} \cmidrule(l{3pt}r{3pt}){12-13}

    Method & bAcc. & F1 & bAcc. & F1 & bAcc. & F1 & bAcc. & F1 & bAcc. & F1 & bAcc. & F1 \\ \midrule

    $k$-NN & 62.0 {$\pm$} 0.0 & 40.3 {$\pm$} 0.0 
    & 76.9 {$\pm$} 0.0 & 71.1 {$\pm$} 0.0     
    & 57.7 {$\pm$} 0.0 & 56.9 {$\pm$} 0.0
    & 81.5 {$\pm$} 0.3 & 47.6 {$\pm$} 0.0
    & \underline{57.6 {$\pm$} 0.1} & \underline{56.9 {$\pm$} 0.0}
    & 67.9 {$\pm$} 0.3 & 53.3 {$\pm$} 0.1 \\
    
    LogReg & 63.5 {$\pm$} 0.0 & 44.2 {$\pm$} 0.0 
    & 80.2 {$\pm$} 0.0 & 76.2 {$\pm$} 0.0     
    & 61.4 {$\pm$} 0.0 & 58.9 {$\pm$} 0.0  
    & 61.6 {$\pm$} 0.0 & 62.2 {$\pm$} 0.0 
    & 50.0 {$\pm$} 0.0 & 47.9 {$\pm$} 0.0 
    & 71.0 {$\pm$} 0.0 & 55.4 {$\pm$} 0.0 \\

    RandomForest & 58.2 {$\pm$} 7.6 & 32.2 {$\pm$} 1.5 
    & \BF{81.7 {$\pm$} 0.1} & 68.4 {$\pm$} 0.7     
    & \underline{64.4 {$\pm$} 0.5} & 42.1 {$\pm$} 1.2  
    & \BF{85.2 {$\pm$} 0.4} & 52.0 {$\pm$} 0.1     
    & 50.0 {$\pm$} 0.0 & 47.9 {$\pm$} 0.0     
    & \BF{76.5 {$\pm$} 0.1} & 46.9 {$\pm$} 0.1 \\
    
    XGBoost & 57.6 {$\pm$} 7.2 & 39.9 {$\pm$} 4.9 
    & \underline{80.5 {$\pm$} 0.2} & 75.8 {$\pm$} 0.4 
    & 63.1 {$\pm$} 0.1 & 61.3 {$\pm$} 0.4
    & 79.9 {$\pm$} 0.1 & \underline{64.3 {$\pm$} 0.1}
    & 50.0 {$\pm$} 0.0 & 47.9 {$\pm$} 0.0
    & 71.5 {$\pm$} 0.1 & 56.2 {$\pm$} 0.1 \\
    
    CatBoost & \underline{65.4 {$\pm$} 0.0} & \underline{51.7 {$\pm$} 0.0}
    & 80.4 {$\pm$} 0.0 & \underline{76.8 {$\pm$} 0.0} 
    & 63.4 {$\pm$} 0.0 & \underline{61.8 {$\pm$} 0.5}  
    & 81.4 {$\pm$} 0.0 & 59.8 {$\pm$} 0.0
    & 50.0 {$\pm$} 0.0 & 47.9 {$\pm$} 0.0
    & 65.0 {$\pm$} 0.0 & \underline{59.3 {$\pm$} 0.0} \\

    \rowcolor{brightgray} + AdapTable  & \BF{65.5 {$\pm$} 0.0} & \BF{65.4 {$\pm$} 0.0}
    & 79.6 {$\pm$} 0.0 & \BF{78.6 {$\pm$} 0.0} 
    & \BF{65.4 {$\pm$} 0.0} & \BF{62.5 {$\pm$} 0.3} 
    & \underline{82.6 {$\pm$} 0.0} & \BF{64.8 {$\pm$} 0.3}
    & \BF{62.8 {$\pm$} 0.4} & \BF{61.7 {$\pm$} 0.3}
    & \underline{74.2 {$\pm$} 0.0} & \BF{62.5 {$\pm$} 0.3} \\ \midrule

    Source & \underline{53.2 {$\pm$} 1.5} & \underline{38.2 {$\pm$} 3.5} 
    & \underline{76.5 {$\pm$} 0.5} & 77.3 {$\pm$} 0.4
    & \underline{61.1 {$\pm$} 0.1} & \underline{60.2 {$\pm$} 0.3}
    & 56.3 {$\pm$} 0.0 & 58.1 {$\pm$} 0.0
    & 50.0 {$\pm$} 0.0 & 47.9 {$\pm$} 0.0
    & 55.2 {$\pm$} 0.0 & 55.5 {$\pm$} 0.0 \\

    PL & 51.8 {$\pm$} 1.0 & 34.9 {$\pm$} 2.3 
    & 75.6 {$\pm$} 0.5 & 76.6 {$\pm$} 0.5 
    & 60.5 {$\pm$} 0.1 & 58.9 {$\pm$} 0.3  
    & 56.3 {$\pm$} 0.0 & 58.0 {$\pm$} 0.1
    & 50.0 {$\pm$} 0.0 & 47.9 {$\pm$} 0.0
    & 55.1 {$\pm$} 0.1 & 55.3 {$\pm$} 0.0 \\
    
    TTT++ & \underline{53.2 {$\pm$} 1.5} & \underline{38.2 {$\pm$} 3.6}
    & 76.8 {$\pm$} 0.5 & \underline{77.6 {$\pm$} 0.2}
    & \underline{61.1 {$\pm$} 0.1} & \underline{60.2 {$\pm$} 0.3}
    & \underline{56.6 {$\pm$} 0.5} & \underline{58.5 {$\pm$} 0.1}
    & 50.0 {$\pm$} 0.0 & 47.9 {$\pm$} 0.0
    & \underline{55.4 {$\pm$} 0.0} & \underline{55.7 {$\pm$} 0.0} \\
    
    TENT & 51.2 {$\pm$} 1.2 & 33.2 {$\pm$} 2.6 
    & 74.0 {$\pm$} 0.6 & 74.9 {$\pm$} 0.6 
    & 60.2 {$\pm$} 0.1 & 58.3 {$\pm$} 0.3
    & 55.1 {$\pm$} 0.1 & 56.3 {$\pm$} 0.1
    & 50.0 {$\pm$} 0.0 & 47.9 {$\pm$} 0.0
    & 55.0 {$\pm$} 0.0 & 55.0 {$\pm$} 0.0 \\

    EATA & \underline{53.2 {$\pm$} 1.5} & \underline{38.2 {$\pm$} 3.6} 
    & \underline{76.5 {$\pm$} 0.5} & 77.3 {$\pm$} 0.4
    & \underline{61.1 {$\pm$} 0.1} & \underline{60.2 {$\pm$} 0.4}
    & 56.3 {$\pm$} 0.0 & 58.1 {$\pm$} 0.0
    & 50.0 {$\pm$} 0.0 & 47.9 {$\pm$} 0.0
    & 55.2 {$\pm$} 0.0 & 55.5 {$\pm$} 0.0 \\

    SAR & 50.0 {$\pm$} 0.0 & 30.1 {$\pm$} 0.0 
    & 62.0 {$\pm$} 1.2 & 59.4 {$\pm$} 1.6 
    & 57.1 {$\pm$} 1.1 & 51.3 {$\pm$} 2.2
    & 51.1 {$\pm$} 0.1 & 49.1 {$\pm$} 0.2
    & 50.0 {$\pm$} 0.0 & 47.9 {$\pm$} 0.0
    & 53.4 {$\pm$} 0.0 & 52.2 {$\pm$} 0.0 \\

    LAME & 50.0 {$\pm$} 0.0 & 30.1 {$\pm$} 0.0 
    & 54.6 {$\pm$} 0.5 & 46.8 {$\pm$} 1.0 
    & 54.9 {$\pm$} 0.5 & 46.9 {$\pm$} 1.0
    & 50.0 {$\pm$} 0.0 & 46.7 {$\pm$} 0.0
    & 50.0 {$\pm$} 0.0 & 47.9 {$\pm$} 0.0
    & 54.8 {$\pm$} 0.1 & 54.8 {$\pm$} 0.2 \\

    \rowcolor{brightgray} AdapTable & \BF{65.8 {$\pm$} 0.6} & \BF{64.5 {$\pm$} 0.3}
    & \BF{78.4 {$\pm$} 0.3} & \BF{78.6 {$\pm$} 0.0} 
    & \BF{61.7 {$\pm$} 0.0} & \BF{61.7 {$\pm$} 0.0}
    & \BF{65.9 {$\pm$} 0.1} & \BF{65.4 {$\pm$} 0.1}
    & \BF{69.2 {$\pm$} 0.1} & \BF{60.9 {$\pm$} 0.3}
    & \BF{70.9 {$\pm$} 0.1} & \BF{68.3 {$\pm$} 0.1} \\
    
    \bottomrule
\end{tabular}
\end{adjustbox}
\end{footnotesize}
\end{small}
\vspace{-.2in}
\end{table*}

%% file: figures/main_common_corruption.tex
\begin{figure*}[!t]
\centering
\includegraphics[width=.9\linewidth]{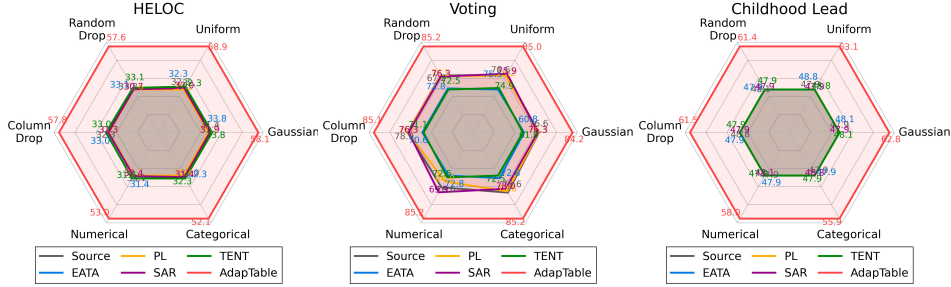}
\caption{
    The average macro F1 score for AdapTable and TTA baselines is reported under six common corruptions using MLP across three datasets within the TableShift~\cite{tableshift} benchmark. % The results are averaged over three random repetitions.
}
\label{fig:main_common_corruption}
\vspace{-.25in}
\end{figure*}

%% file: figures/calibration_efficacy.tex
\begin{figure}[!ht]
\centering
\includegraphics[width=.65\linewidth]{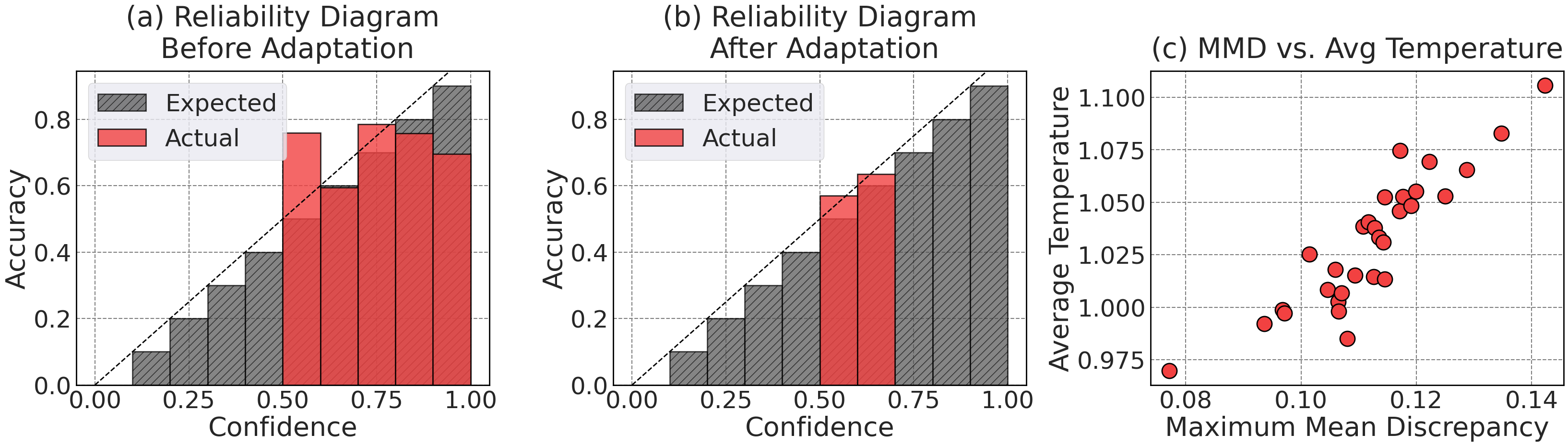}
\vspace{-.1in}
\caption{
    Ablation study on the shift-aware uncertainty calibrator using MLP for the HELOC dataset. (a) and (b) show reliability diagrams before and after calibration, while (c) depicts the average temperature relative to the maximum mean discrepancy (MMD) between the training set and the sampled test sets.
}
\label{fig:calibration_efficacy}
\vspace{-.2in}
\end{figure}

%% file: tables/ablation_uncertainty_calibrator.tex
\begin{table}[!ht]
\centering
\caption{
    Ablation study comparing the shift-aware uncertainty calibrator with classical methods---Platt scaling (PS) and isotonic regression (IR). The results are averaged over three random repetitions.
}
\label{table:ablation_uncertainty_calibrator}
\begin{adjustbox}{width=.4\linewidth}
\begin{tabular}{lcccccc}
    \toprule
    
    Method & HELOC & Voting & Hospital Readmission \\ \midrule
    
    Source & 38.2 $\pm$ 3.5 & \underline{77.3 $\pm$ 0.4} & \underline{60.2 $\pm$ 0.3} \\
    
    PS & \underline{61.6 $\pm$ 1.3} & 73.3 $\pm$ 0.2 & 59.4 $\pm$ 0.3 \\
    
    IR & 61.3 $\pm$ 1.7 & 74.3 $\pm$ 0.2 & 58.0 $\pm$ 0.4 \\
    
    \rowcolor{brightgray} AdapTable & \textbf{64.5 $\pm$ 0.6} & \textbf{78.6 $\pm$ 0.0} & \textbf{61.7 $\pm$ 0.0} \\
    
    \bottomrule
\end{tabular}
\end{adjustbox}
\vspace{-.1in}
\end{table}

%% file: figures/js_div.tex
\begin{figure}[!ht]
\centering
\vspace{-.1in}
\includegraphics[width=.65\linewidth]{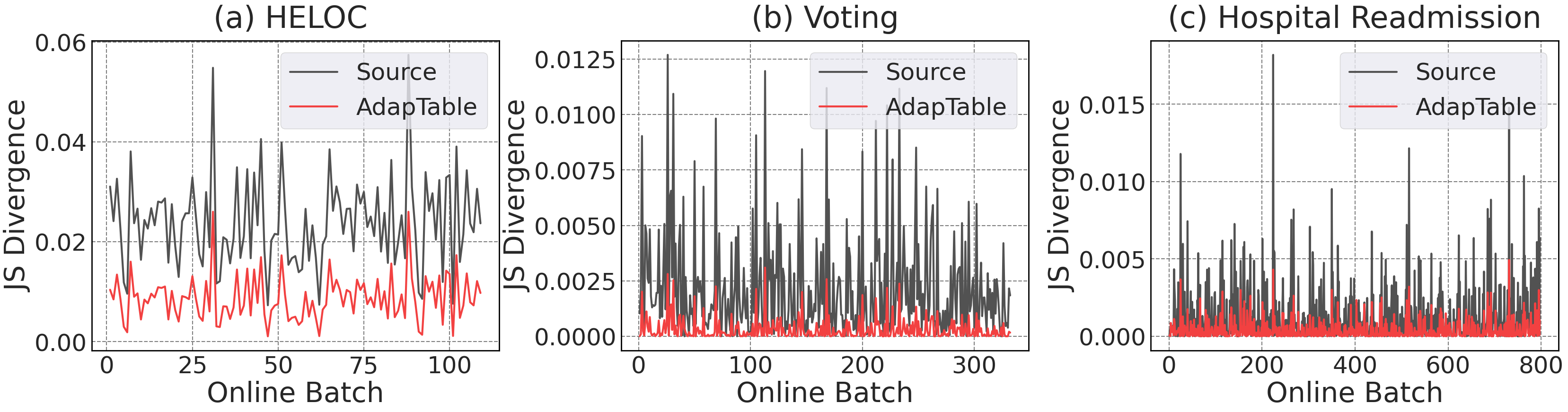}
\vspace{-.1in}
\caption{
    Jensen-Shannon (JS) Divergence of the estimated target label distribution before and after applying the label distribution handler using MLP on three datasets. The x-axis indicates the online batch index, and the y-axis shows the per-batch JS divergence from the ground truth labels.
}
\label{fig:js_div}
\vspace{-.2in}
\end{figure}

%% file: tables/label_distribution_shift.tex
\begin{table}[!ht]
\vspace{-.15in}
\caption{
    The average macro F1 score (\%) with standard errors for TTA baselines is reported using MLP across three datasets with 1) class imbalance and 2) temporal correlation from the TableShift benchmark. The results are averaged over three random repetitions.
}
\label{table:label_distribution_shift}
\centering
\setlength{\columnsep}{1pt}
\begin{adjustbox}{width=.55\linewidth}
\begin{tabular}{lcccccc}
    \toprule

    & \multicolumn{3}{c}{Class Imbalance} & \multicolumn{3}{c}{{Temporal Correlation}} \\ \cmidrule(l{3pt}r{3pt}){2-4} \cmidrule(l{3pt}r{3pt}){5-7} \rule{0pt}{2.2ex}
    
    Method & HELOC & Voting & Childhood Lead & HELOC & Voting & Childhood Lead \\ \midrule
    
    Source & \underline{32.5 {$\pm$} 3.5} & 52.3 {$\pm$} 4.9 & 36.7 {$\pm$} 6.5 & 31.6 {$\pm$} 0.3 & \underline{62.2 {$\pm$} 0.1} & 35.1 {$\pm$} 0.2 \\
    
    PL & 32.0 {$\pm$} 3.6 & 52.1 {$\pm$} 4.9 & 36.7 {$\pm$} 6.5 & 30.9 {$\pm$} 0.2 & 54.9 {$\pm$} 0.1 & 35.1 {$\pm$} 0.2 \\
    
    TENT & \underline{32.5 {$\pm$} 3.5} & 52.3 {$\pm$} 4.9 & 36.7 {$\pm$} 6.5 & 31.6 {$\pm$} 0.3 & 55.7 {$\pm$} 0.1 & 35.1 {$\pm$} 0.2 \\
    
    EATA & \underline{32.5 {$\pm$} 3.5} & 52.3 {$\pm$} 4.9 & 36.7 {$\pm$} 6.5 & 31.6 {$\pm$} 0.3 & 55.7 {$\pm$} 0.1 & 35.1 {$\pm$} 0.2 \\
    
    SAR & 31.8 {$\pm$} 3.5 & 57.1 {$\pm$} 5.3 & 36.7 {$\pm$} 6.5 & \underline{32.0 {$\pm$} 0.2} & 54.4 {$\pm$} 0.5 & 35.1 {$\pm$} 0.2 \\
    
    LAME & 29.9 {$\pm$} 3.5 & \underline{58.7 {$\pm$} 4.0} & 36.7 {$\pm$} 6.5 & 29.0 {$\pm$} 0.1 & 38.0 {$\pm$} 0.4 & 35.1 {$\pm$} 0.2 \\
    
    \rowcolor{brightgray} AdapTable & \textbf{59.7 {$\pm$} 0.8} & \textbf{62.0 {$\pm$} 4.6} & \textbf{63.9 {$\pm$} 1.0} & \textbf{56.1 {$\pm$} 0.3} & \textbf{64.5 {$\pm$} 0.0} & \textbf{64.8 {$\pm$} 0.3} \\

    \bottomrule
\end{tabular}
\end{adjustbox}
\vspace{-.2in}
\end{table}

%% file: sections/6_conclusion.tex
\section{Conclusion}

In this paper, we have introduced AdapTable, a test-time adaptation framework tailored for tabular data. AdapTable overcomes the limitations of previous methods, which fail to address label distribution shifts, and lack versatility across architectures. Our approach, combined with a shift-aware uncertainty calibrator that enhances calibration via modeling column shifts, and a label distribution handler that adjusts the output distribution based on real-time estimates of the current batch's label distribution. Extensive experiments show that AdapTable achieves state-of-the-art performance across various datasets and architectures, effectively managing both natural distribution shifts and common corruptions.

%% file: sections/7_acknowledgments.tex
\section*{Acknowledgments}

This work was supported by the Institute for Information \& Communications Technology Planning \& Evaluation (IITP) grant funded by the Korea government (MSIP) (No. 2019-0-00075, Artificial Intelligence Graduate School Program (KAIST)).

%% file: sections/8_appendix.tex
\clearpage
\appendix
\centerline{\Large\bf Appendix}
%%%%%%%%%%%%%%%%%%%%%%%%%%%%%%%%%%%%%%%%%%%%%%%%%%%%%%%%%%%%

%%%%%%%%%%%%%%%%%%%%%%%%%%%%%%%%%%%%%%%%%%%%%%%%%%%%%%%%%%%%
\input{sections/5_related_work}
%%%%%%%%%%%%%%%%%%%%%%%%%%%%%%%%%%%%%%%%%%%%%%%%%%%%%%%%%%%%

%%%%%%%%%%%%%%%%%%%%%%%%%%%%%%%%%%%%%%%%%%%%%%%%%%%%%%%%%%%%
\section{Detailed Algorithm of AdapTable} \label{sec:adaptable_detail}

\paragraph{Post-training shift-aware uncertainty calibrator.}
Given a pre-trained tabular classifier $ f_{\theta}: \mathbb{R}^{D} \rightarrow \mathbb{R}^{C} $ on the source domain $ \mathcal{D}_s = \{(\mathbf{x}_i^s, y_i^s )\}_{i} $, we introduce a post-training phase for a shift-aware uncertainty calibrator $ g_{\phi}: \mathbb{R}^{C} \times \mathbb{R}^{D \times N} \rightarrow \mathbb{R}^{+} $. This calibrator is trained after the initial training of $ f_{\theta} $ using the same training dataset $ \mathcal{D}_s $. For a given training batch $ \{(\mathbf{x}_i^{s}, y_i^{s})\}_{i=1}^{N} $, we compute the shift trend $ \mathbf{s}^{s} = (\mathbf{s}_{u}^{s})_{u=1}^{D} $ for a specific column index $ u $ as follows:
\begin{equation*}
    \mathbf{s}_{u}^{s} = \big( \mathbf{x}^{s}_{iu} - \frac{1}{|\mathcal{D}_s|} \sum_{i'=1}^{|\mathcal{D}_s|} \mathbf{x}_{i'u}^s \big)_{i=1}^{N},
\end{equation*}
where we add a linear layer to $ \mathbf{s}_u^{s} $ for categorical column $ u $ to transform it into a one-dimensional representation, ensuring alignment with the numerical columns. Using $\mathbf{s}^{s}$, we construct a shift trend graph, where each node $u$ represents a column, and edges capture the relationships between columns. The node features are given by $\mathbf{s}_u^t$, and the graph is connected using an all-ones adjacency matrix. A graph neural network (GNN) is applied to this graph, facilitating the exchange of shift trends between columns through message passing, which generates a contextualized column-wise representation $\vh_u^s$. These representations are averaged to form a global feature representation $\vh^{s} = \frac{1}{D}\sum_{u=1}^{D}{ \vh_u^s }$, which is then concatenated with the initial model prediction $f_{\theta}(\mathbf{x}_i^s)$ to produce the final output temperature $T_i$. With the calibrated probability $ p_i = \text{softmax}\big(f_{\theta}(\mathbf{x}_i^{s}) / T_i\big) $, with the per-sample temperature $ T_i $ calculated above, we define the most plausible and second plausible class indices $ j^{*} $ and $ j^{**} $ as follows:
\begin{equation*}
    j^{*} = \argmax_{j \in \mathcal{Y}} p_{ij} \quad \text{and} \quad j^{**} = \argmax_{j \in \mathcal{Y}, j \ne j^{*}} p_{ij}.
\end{equation*}
The focal loss $ \mathcal{L}_{\text{FL}} $~\citep{lin2017focal} and the calibration loss $ \mathcal{L}_{\text{CAL}} $~\citep{wang2021confident} are used to train the shift-aware uncertainty calibrator $ g_{\phi} $, defined as:
\begin{align}
    \mathcal{L}_{\text{FL}}(\mathbf{x}_i^{s}, y_i^s) &= \sum_{j=1}^{C} \mathbbold{1}_{\{ y_i^s \}}(j) (1 - p_{ij})^{\gamma} \log{p_{ij}}, \\
    \mathcal{L}_{\text{CAL}}(\mathbf{x}_i^{s}, y_i^s) &= \mathbbold{1}_{\{ y_{i}^{s} \}}(j^{*}) (1 - p_{ij^{*}} + p_{ij^{**}}) + \mathbbold{1}_{\mathcal{Y} \backslash \{ y_{i}^{s} \}}(j^{*}) (p_{ij^{*}} - p_{ij^{**}}),
\end{align}
where $ \mathbbold{1}_{\mA}(x) $ is an indicator function:
\begin{equation*}
    \mathbbold{1}_{\mA}(x) =  
    \begin{cases}
    1 & \text{if } x \in \mA \\
    0 & \text{otherwise}.
    \end{cases}
\end{equation*}
$ \mathcal{L}_{\text{FL}} $ addresses class imbalance by reducing the impact of easily classified examples, while $ \mathcal{L}_{\text{CAL}} $ penalizes the gap between $ p_{ij^{*}} $ and $ p_{ij^{**}} $ for correct predictions, encouraging them to converge for incorrect predictions. For all experiments, we set $\gamma = 2$ and $\lambda_{\text{CAL}} = 0.1$.

\input{algorithms/adaptable}

\paragraph{Label distribution handler.}
During the test phase after post-training $ g_{\phi} $, we introduce a label distribution handler using an estimator $ \bar{p}_{i}(y|\mathbf{x}_{i}^{t}) $, defined as:
\begin{equation*}
    \bar{p}_{i}(y|\mathbf{x}_{i}^{t}) = \frac{
        \tilde{p}_t(y|\mathbf{x}_{i}^{t}) + \text{norm}\big(\tilde{p}_t(y|\mathbf{x}_{i}^{t}) p_t(y) / p_s(y) \big)
    }{2},
\end{equation*}
where $ \tilde{p}_t(y|\mathbf{x}_{i}^{t}) $ represents the calibrated prediction. This approach enhances uncertainty quantification and combines the calibrated estimation with the distributionally aligned prediction for more robust estimation. To compute $ \tilde{p}_t(y|\mathbf{x}_{i}^{t}) $, we perform a two-stage uncertainty calibration. Specifically, for a given test batch $ \{\mathbf{x}_{i}^{t} \}_{i=1}^{N} $, we calculate the shift trend $ \mathbf{s}^t = (\mathbf{s}_u^t)_{u=1}^{D} \in \mathbb{R}^{D \times N} $ as:
\begin{equation*}
    \mathbf{s}_{u}^{t} = \big( \mathbf{x}^{t}_{iu} - \frac{1}{|\mathcal{D}_s|} \sum_{i'=1}^{|\mathcal{D}_s|} \mathbf{x}_{i'u}^s \big)_{i=1}^{N} \in \mathbb{R}^{N}.
\end{equation*}
Then, a per-sample temperature $ T_i = g_{\phi}(f_{\theta}(\mathbf{x}_i^t), \mathbf{s}^t) $, which was defined in Equation~\ref{eqn:shift_trend} is computed. The uncertainty $ \delta_{i} $ of $ f_{\theta}(\mathbf{x}_i^t) $ is defined as the reciprocal of the margin of the calibrated probability distribution $ \text{softmax}(f_{\theta}(\mathbf{x}_i^t) / T_i) $:
\begin{equation*}
    \delta_{i} = \frac{1}{\text{softmax}\big(f_{\theta}(\mathbf{x}_i^{t}) / T_i\big)_{j^{*}} - { \text{softmax}\big(f_{\theta}(\mathbf{x}_i^{t}) / T_i \big) }_{j^{**}}},
\end{equation*}
where $ j^{*} $ and $ j^{**} $ are the most plausible and second plausible class indices:
\begin{equation*}
    j^{*} = \argmax_{j \in \mathcal{Y}} f_{\theta}(\mathbf{x}_i^{t})_{j} \quad \text{and} \quad j^{**} = \argmax_{j \in \mathcal{Y}, j \ne j^{*}} f_{\theta}(\mathbf{x}_i^{t})_{j}.
\end{equation*}
Based on $\delta_{i}$, the recalibrated temperature $\tilde{T}_i $is applied:
\begin{equation*}
    \tilde{T}_i =
    \begin{cases}
        T & \text{if } \delta_{i} \ge Q\big(\{\delta_{i'}\}_{i'=1}^{N}, q_{\text{high}}\big) \\
        1 / T & \text{if } \delta_{i} \le Q\big(\{\delta_{i'}\}_{i'=1}^{N}, q_{\text{low}}\big) \\
        1 & \text{otherwise},
    \end{cases}
\end{equation*}
where $ T = 1.5 \rho / (\rho - 1 + 10^{-6}) $ with $\rho = \max_{j} p_{s}(y)_{j} / \min_{j} p_{s}(y)_{j}$, and $ q_{\text{low}} $ and $ q_{\text{high}} $ are the low and high uncertainty quantiles, respectively. The target label distribution $ p_t(y) $ is then estimated using the following formula:
\begin{equation*}
    p_t(y) = (1 - \alpha) \cdot \frac{1}{N} \sum_{i=1}^{N} p^{\text{de}}_{t}(y|\mathbf{x}_{i}^{t}) + \alpha \cdot p^{\text{oe}}_t(y),
\end{equation*}
where $ p^{\text{de}}_{t}(y|\mathbf{x}_{i}^{t}) = \text{norm}\big(p_t(y|\mathbf{x}_{i}^{t}) / p_s(y)\big) $ serves as a debiased target label estimator, deviating from the source label distribution $ p_s(y) $. The online target label estimator $ p_t^{\text{oe}}(y) $ is initialized with a uniform distribution and updated with each new batch as follows:
\begin{equation*}
    p^{\text{oe}}_{t}(y) = (1 - \alpha) \cdot \frac{1}{N} \sum_{i=1}^{N} \bar{p}_{t}(y|\mathbf{x}_{i}^{t}) + \alpha \cdot p^{\text{oe}}_{t}(y),
\end{equation*}
where $ \alpha $ is a smoothing factor. This update process leverages information from the current batch to refine the target label distribution estimation over time. The overall procedure of the proposed AdapTable method is summarized in Algorithm~\ref{alg:adaptable}.

%%%%%%%%%%%%%%%%%%%%%%%%%%%%%%%%%%%%%%%%%%%%%%%%%%%%%%%%%%%%

%%%%%%%%%%%%%%%%%%%%%%%%%%%%%%%%%%%%%%%%%%%%%%%%%%%%%%%%%%%%
\section{Proof of Theorem~\ref{thm:main}} \label{sec:adaptable_proof}
% We regret to inform the reviewers that we have identified a minor error in Theorem 3.1. The revised version is presented as Theorem~\ref{thm:main_fixed}. The corrected version will also be incorporated into our final draft.
Let's first define the balanced source error $BSE(\hat{Y})$ on the source dataset and the conditional error gap $\Delta_{CE}(\hat{Y})$ between $\mathbb{P}(\hat{Y} \ne Y|X_s)$ and $\mathbb{P}(\hat{Y} \ne Y|X_t)$ as follows:
\begin{align}
BSE(\hat{Y}) &= \max_{i \in \mathcal{Y}} \mathbb{P}(\hat{Y} \ne i | Y = i, X_s), \\
\Delta_{CE}(\hat{Y}) &= \max_{i \ne i' \in \mathcal{Y}} \Big| \mathbb{P}(\hat{Y} = i | Y = i', X_s) - \mathbb{P}(\hat{Y} = i | Y = i', X_t) \Big|.
\end{align}

\begin{defn} \label{defn:gls}
(Generalized Label Shift in~\citet{tachet2020domain}). % Let $\mathbb{P}(Y = i | X_s) = p_s(y)_i$ and $\mathbb{P}(Y = i | X_t) = p_t(y)_i$, respectively.
Both input covariate distribution $\mathbb{P}(X_s) \ne \mathbb{P}(X_t)$ and output label distribution $\mathbb{P}(Y | X_s) \ne \mathbb{P}(Y | X_t)$ change. Yet, there exists a hidden representation $H = g^{*}(X)$ such that the conditional distribution of $H$ given $Y$ remains the same across both domains, \ie, $\forall i \in \mathcal{Y}$,
\begin{equation}
    \mathbb{P}(H|Y = i, X_s) = \mathbb{P}(H|Y = i, X_t).
\end{equation}
\end{defn}

% \begin{thm} \label{thm:main_fixed}
% Let $\hat{Y}|X$ and $\hat{Y}_{o}|X$ be defined as follows:
% \begin{align}
%     \hat{Y}|X &= \{ \argmax_{j \in \mathcal{Y}}{f_{\theta}(\mathbf{x})_{j}}|\mathbf{x} \in X \}, \\
%     \hat{Y}_{o}|X &= \{ \argmax_{j \in \mathcal{Y}}{f_{\theta}(\mathbf{x})_{j} + \log p_{t}^{oe}(y)_j}|\mathbf{x} \in X\}.
% \end{align}
% Given the error $\epsilon(\hat{Y} | X) = \mathbb{P}(\hat{Y} \ne Y | X)$, with true labels $Y$ of inputs $X$, the error gap $| \epsilon(\hat{Y} | X_s) - \epsilon(\hat{Y}_{o}|X_t) |$ is upper bounded by
% \begin{equation}
% K_1 \Big\| 1 - \frac{p_t^{oe}(y)}{p_t(y)} \Big\|_{1} BSE(\hat{Y}) + K_2 \Delta_{CE}(\hat{Y}),
% \end{equation}
% where $K_1$ and $K_2$ are constants related to $p_t(y)$, and $p_s(y)$, respectively.
% \end{thm}
% \paragraph{Proof of Theorem~\ref{thm:main}.}
% Proof of Theorem~\ref{thm:main}.
\begin{proof}
We start by applying the law of total probability and triangle inequality to derive the following inequality:
\begin{align} \label{eqn:basic}
\begin{split}
    &\Big| \epsilon(\hat{Y} | X_s) - \epsilon(\hat{Y}_{o}|X_t) \Big| \\
    &=\Big| \mathbb{P}(\hat{Y} \ne Y | X_s) - \mathbb{P}(\hat{Y}_{o} \ne Y | X_t) \Big| \\
    &=\Big| \sum_{i \ne i'} \mathbb{P}(\hat{Y} = i, Y = i' | X_s) - \sum_{i \ne i'} \mathbb{P}(\hat{Y}_{o} = i, Y = i' | X_t) \Big| \\
    &=\Big| \sum_{i \ne i'} \mathbb{P}(Y = i' | X_s) \mathbb{P}(\hat{Y} = i | Y = i', X_s) - \sum_{i \ne i'} \mathbb{P}(Y = i' | X_t) \mathbb{P}(\hat{Y}_{o} = i | Y = i', X_t) \Big| \\
    &\le \sum_{i \ne i'} \Big| \mathbb{P}(Y = i' | X_s) \mathbb{P}(\hat{Y} = i | Y = i', X_s) - \mathbb{P}(Y = i' | X_t) \mathbb{P}(\hat{Y}_{o} = i | Y = i', X_t) \Big|.
\end{split}
\end{align}
According to Equation 8 in \cite{menon2020long}, $\hat{Y}_{o}$ satisfies the following condition under generalized label shift condition in Definition~\ref{defn:gls}:
\begin{align} \label{eqn:long}
    \mathbb{P}(\hat{Y}_{o} = i | H, X_t) = \frac{p_t^{oe}(y)_{i}}{\mathbb{P}(Y = i | X_s)} \mathbb{P}(\hat{Y} = i | H, X_t).
\end{align}
By multiplying both sides of Equation~\ref{eqn:long} by $\mathbb{P}(H | Y, X_t)$, we obtain:
\begin{align} \label{eqn:cond}
\begin{split}    
    \mathbb{P}(\hat{Y}_{o} = i | H, X_t) \mathbb{P}(H | Y, X_t) &= \frac{p_t^{oe}(y)_{i}}{\mathbb{P}(Y = i | X_s)} \mathbb{P}(\hat{Y} = i | H, X_t) \mathbb{P}(H | Y, X_t) \\
    \mathbb{P}(\hat{Y}_{o} = i | Y, X_t) &= \frac{p_t^{oe}(y)_{i}}{\mathbb{P}(Y = i | X_s)} \mathbb{P}(\hat{Y} = i | Y, X_t).
\end{split}
\end{align}
Next, by substituting Equation~\ref{eqn:cond} into Equation~\ref{eqn:basic}, and letting $Y = i'$, we have:
\begin{align} \label{eqn:interm_upper_bound}
\begin{split}    
    &\Big| \epsilon(\hat{Y} | X_s) - \epsilon(\hat{Y}_{o}|X_t) \Big| \\
    &\le \sum_{i \ne i'} \Big| \mathbb{P}(Y = i' | X_s) \mathbb{P}(\hat{Y} = i | Y = i', X_s) - \mathbb{P}(Y = i' | X_t) \frac{p_t^{oe}(y)_{i}}{\mathbb{P}(Y = i | X_s)} \mathbb{P}(\hat{Y} = i | Y = i', X_t) \Big|.
\end{split}
\end{align}
Using Lemma A.2 from \cite{tachet2020domain}, we can further estimate the upper bound of Equation~\ref{eqn:interm_upper_bound} as follows:
\begin{align}
\begin{split}
    &\Big| \epsilon(\hat{Y} | X_s) - \epsilon(\hat{Y}_{o} | X_t) \Big| \\
    &\le \sum_{i \ne i'} \mathbb{P}(Y = i' | X_t) \Bigg| 1 - \frac{p_t^{oe}(y)_{i}}{\mathbb{P}(Y = i | X_s)} \Bigg| \left( \alpha_{i'} \mathbb{P}(\hat{Y} = i | Y = i', X_s) + \beta_{i'} \mathbb{P}(\hat{Y} = i | Y = i', X_t) \right) \\
    &\quad + \mathbb{P}(Y = i' | X_s) \Delta_{CE}(\hat{Y}) + \mathbb{P}(Y = i' | X_t) \frac{p_t^{oe}(y)_{i}}{\mathbb{P}(Y = i | X_s)} \Delta_{CE}(\hat{Y}) \\
    &\stackrel{\mathclap{(i)}}{\le} \sum_{i \ne i'} \mathbb{P}(Y = i' | X_t) \Bigg| 1 - \frac{p_t^{oe}(y)_{i}}{\mathbb{P}(Y = i | X_s)} \Bigg| \left( \alpha_{i'} \mathbb{P}(\hat{Y} = i | Y = i', X_s) + \beta_{i'} \mathbb{P}(\hat{Y} = i | Y = i', X_t) \right) \\
    &\quad + (C - 1) \Delta_{CE}(\hat{Y}) + \left( \sum_{i \ne i'} \frac{\mathbb{P}(Y = i' | X_t)}{\mathbb{P}(Y = i | X_s)} \right) \left( \sum_{i \ne i'} p_t^{oe}(y)_{i} \right) \Delta_{CE}(\hat{Y}) \\
    &\le \sum_{i \ne i'} \mathbb{P}(Y = i' | X_t) \Bigg| 1 - \frac{p_t^{oe}(y)_{i}}{\mathbb{P}(Y = i | X_s)} \Bigg| \left( \alpha_{i'} \mathbb{P}(\hat{Y} = i | Y = i', X_s) + \beta_{i'} \mathbb{P}(\hat{Y} = i | Y = i', X_t) \right) \\
    &\quad + (C - 1) \Delta_{CE}(\hat{Y}) + \frac{(C - 1)^2}{\min_{i \in \mathcal{Y}} \mathbb{P}(Y = i | X_s)} \Delta_{CE}(\hat{Y}),
\end{split}
\end{align}
where $\alpha_{i'}, \beta_{i'} \ge 0$ and $\alpha_{i'} + \beta_{i'} = 1$, $(i)$ holds by H\"older's inequality.
By letting $\alpha_{i'} = 1$ and $\beta_{i'} = 0$ for all $i' \in \mathcal{Y}$, and defining $K_1$ and $K_2$ as:
\begin{align*}
K_1 &= C (C-1)^2 \max_{i \in \mathcal{Y}} \mathbb{P}(Y = i | X_t), \\
K_2 &= (C - 1) + \frac{(C - 1)^2}{\min_{i \in \mathcal{Y}} \mathbb{P}(Y = i | X_s)},
\end{align*}
we finally get:
\begin{align}
\begin{split}
    &\Big| \epsilon(\hat{Y} | X_s) - \epsilon(\hat{Y}_{o}|X_t) \Big| \\
    &\le \sum_{i \ne i'} \mathbb{P}(Y = i' | X_t) \Bigg| 1 - \frac{p_t^{oe}(y)_{i}}{\mathbb{P}(Y = i | X_s)} \Bigg| \mathbb{P}(\hat{Y} = i | Y = i', X_s) + K_2 \Delta_{CE}(\hat{Y}) \\
    &\le \max_{i' \in \mathcal{Y}} \mathbb{P}(Y = i' | X_t) \sum_{i \ne i'} \Bigg| 1 - \frac{p_t^{oe}(y)_i}{\mathbb{P}(Y = i | X_s)} \Bigg| \mathbb{P}(\hat{Y} = i | Y = i', X_s) + K_2 \Delta_{CE}(\hat{Y}) \\
    &\stackrel{\mathclap{(i)}}{\le} \max_{i' \in \mathcal{Y}} \mathbb{P}(Y = i' | X_t) \left( \sum_{i \ne i'} \Bigg| 1 - \frac{p_t^{oe}(y)_i}{\mathbb{P}(Y = i | X_s)} \Bigg| \right) \left( \sum_{i \ne i'} \mathbb{P}(\hat{Y} = i | Y = i', X_s) \right) + K_2 \Delta_{CE}(\hat{Y}) \\
    &\stackrel{\mathclap{(ii)}}{\le} \max_{i' \in \mathcal{Y}} \mathbb{P}(Y = i' | X_t) (C - 1) \sum_{i=1}^{C} \Big| 1 - \frac{p_t^{oe}(y)_i}{\mathbb{P}(Y = i | X_s)} \Big| C (C - 1) BSE(\hat{Y}) + K_2 \Delta_{CE}(\hat{Y}) \\
    &= \max_{i' \in \mathcal{Y}} \mathbb{P}(Y = i' | X_t) C (C-1)^2 \Big\| 1 - \frac{p_t^{oe}(y)}{p_t(y)} \Big\|_{1} BSE(\hat{Y}) + K_2 \Delta_{CE}(\hat{Y}) \\
    &\stackrel{\mathclap{(iii)}}{=} K_1 \Big\| 1 - \frac{p_t^{oe}(y)}{p_t(y)} \Big\|_{1} BSE(\hat{Y}) + K_2 \Delta_{CE}(\hat{Y}),
\end{split}
\end{align}
where $(i)$ holds by H\"older's inequality, $(ii)$ holds by the definition of $BSE(\hat{Y})$, and $(iii)$ holds by the definition of $K_1$.
\end{proof}

We observe that in practice, using $\hat{Y}_{o}|X = \{ \argmax_{j \in \mathcal{Y}}{f_{\theta}(\mathbf{x})_{j} + \log p_{t}^{oe}(y)_j}|\mathbf{x} \in X\}$ can result in performance degradation due to an error accumulation in $p_t^{oe}(y)$. However, our approach, which integrates a two-stage uncertainty calibration with $g_{\phi}$ and a debiased target label estimator $p_{t}^{de}(y)$, demonstrates empirical efficacy across various experiments.

%%%%%%%%%%%%%%%%%%%%%%%%%%%%%%%%%%%%%%%%%%%%%%%%%%%%%%%%%%%%
\section{Dataset Descriptions} \label{sec:dataset_details}

\subsection{Natural Distibution Shifts} \label{subsec:datasets}

In our experiments, we verify our method across six different datasets---HELOC, Voting, Hospital Readmission, ICU Mortality, Childhood Lead, and Diabetes---within the Tableshift Benchmark~\citep{tableshift}, all of which include natural distribution shifts between training and test data.
% Among them, three datasets ({HELOC}, {Voting}, and {Hospital Readmission}) include natural distribution shifts between training and test data, while the other ones \ch{({CMC}, {MFEAT-PIXEL}, and {DNA})} does not have such shifts, and thus we inject noises (Section~\ref{subsec:common_corruptions}) on them to mimic plausible distribution shift scenarios.
% Each dataset is partitioned as follows: 60\% for training, 20\% for validation, and 20\% for testing.
For all datasets, the numerical features are normalized---subtraction of mean and division by standard deviation, while categorical features are one-hot encoded. We find that different encoding types do not play a significant role in terms of accuracy, as noted in~\citet{tree}. Detailed statistics specifications of each dataset are listed in Table~\ref{table:dataset_specification}.

\begin{itemize}
    \item \textbf{HELOC:} This task predicts Home Equity Line of Credit (HELOC)~\citep{heloc} repayment using FICO data~\cite{FICO}, focusing on shifts in third-party risk estimates. The dataset includes 10,459 observations, and a distribution shift occurs by using the 'External Risk Estimate' as a domain split. Estimates above 63 are used for training, while those 63 or below are held out for testing, illustrating potential biases in credit assessments.

    \item \textbf{Voting:} Using ANES~\citep{anes} data, this task predicts U.S. presidential election voting behavior with 8,280 observations. Distribution shift is introduced by splitting the data based on geographic region, with the southern U.S. serving as the out-of-domain region. This simulates how voter behavior predictions might vary when polling data is collected in one region and used to predict outcomes in another.
    
    \item \textbf{Hospital Readmission:} Hospital Readmission~\citep{diabetes_readmission} predicts 30-day readmission of diabetic patients using data from 130 U.S. hospitals over 10 years. The distribution shift occurs by splitting the data based on admission source, with emergency room admissions held out as the target domain. This tests how well models trained on other sources perform when applied to patients admitted through the emergency room.

    \item \textbf{ICU Mortality:} The task predicts ICU patient mortality using MIMIC-iii data~\citep{johnson2016mimic}, focusing on shifts related to insurance type. The dataset includes 23,944 observations, and a distribution shift is created by excluding Medicare and Medicaid patients from the training set, designating them as the target domain. This highlights how insurance type can affect mortality predictions.

    \item \textbf{Childhood Lead:} This task predicts elevated blood lead levels in children using NHANES data~\citep{centers2003national}, with 27,499 observations. A distribution shift is introduced by splitting the data based on poverty using the poverty-income ratio (PIR) as a threshold. Those with a PIR of 1.3 or lower are held out for testing, simulating risk assessment in lower-income households.

    \item \textbf{Diabetes:} This task predicts diabetes using BRFSS data~\citep{american2018economic}, focusing on racial shifts across 1.4 million observations. Distribution shift occurs by focusing on the differences in diabetes risk between racial and ethnic groups, particularly highlighting the higher risk faced by non-white groups compared to White non-Hispanic individuals.

\end{itemize}

\input{tables/dataset_details}

\subsection{Common Corruptions} \label{subsec:common_corruptions}

Let $\mathbf{x}_{i}^{t} = (\mathbf{x}_{ij}^{t})_{j=1}^{D} \in \mathbb{R}^{D}$ be the $i$-th row of a table with $D$ columns in the test data. We define $\bar{\mathbf{x}}_{j}^{s}$ as a random variable that follows the empirical marginal distribution of the $j$-th column in the training set $\mathcal{D}_s$, given by:
\begin{equation*}
\mathbb{P}(\bar{\mathbf{x}}_{j}^{s} = k) = \frac{1}{|\mathcal{D}_s|} \sum_{i=1}^{|\mathcal{D}_s|} \mathbbold{1}_{\{k\}}(\mathbf{x}_{ij}^{s}),
\end{equation*}
where $k \in \mathbb{R}$. Additionally, let $\mu_j^{s} = \mathbb{E}[\bar{\mathbf{x}}_{j}^{s}]$ and $\sigma_j^{s} = \sqrt{\text{Var}(\bar{\mathbf{x}}_{j}^{s})}$ be the mean and standard deviation of the random variable $\bar{\mathbf{x}}_{j}^{s}$, respectively. To effectively simulate natural distribution shifts that commonly occur beyond label distribution shifts, we introduce six types of corruptions---Gaussian noise (\textbf{Gaussian}), uniform noise (\textbf{Uniform}), random missing values (\textbf{Random Drop}), common column missing across all test data (\textbf{Column Drop}), important numerical column shift (\textbf{Numerical}), and important categorical column shift (\textbf{Categorical})---as follows:
\begin{itemize}
    \item \textbf{Gaussian:} For $\mathbf{x}_{ij}^{t}$, Gaussian noise $z \sim \mathcal{N}(0, 0.1^2)$ is independently injected as:
    \begin{equation*}
        \mathbf{x}_{ij}^{t} \gets \mathbf{x}_{ij}^{t} + z \cdot \sigma_{j}^{s}.
    \end{equation*}

    \item \textbf{Uniform:} For the $\mathbf{x}_{ij}^{t}$, uniform noise $u \sim \mathcal{U}(-0.1, 0.1)$ is independently injected as:
    \begin{equation*}
        \mathbf{x}_{ij}^{t} \gets \mathbf{x}_{ij}^{t} + u \cdot \sigma_{j}^{s}.
    \end{equation*}
    
    \item \textbf{Random Drop:} For each column $\mathbf{x}_{ij}^{t}$, a random mask $m_{ij} \sim \text{Bernoulli}(0.2)$ is applied, and the feature is replaced by a random sample $\bar{\mathbf{x}}_{j}^{s}$ drawn from the empirical marginal distribution of the $j$-th column of the training set:
    \begin{equation*}
        \mathbf{x}_{ij}^{t} \gets (1 - m_{ij}) \cdot \mathbf{x}_{ij}^{t} + m_{ij} \cdot \bar{\mathbf{x}}_{j}^{s}.
    \end{equation*}
    
    \item \textbf{Column Drop:} For each column $\mathbf{x}_{ij}^{t}$, a random mask $m_{j} \sim \text{Bernoulli}(0.2)$ is applied, and the feature is replaced by a random sample $\bar{\mathbf{x}}_{j}^{s}$ as follows:
    \begin{equation*}
        \mathbf{x}_{ij}^{t} \gets (1 - m_{j}) \cdot \mathbf{x}_{ij}^{t} + m_{j} \cdot \bar{\mathbf{x}}_{j}^{s}.
    \end{equation*}
    Unlike random drop corruption, where the mask $m_{ij}$ is resampled for each $j$-th column of the $i$-th test instance $\mathbf{x}_{ij}^{t}$, a single random mask $m_{j}$ is sampled for each $j$-th column and applied uniformly across all test data.

    \item \textbf{Numerical:} Important numerical column shift simulates natural domain shifts where the test distribution of the most important numerical column deviates significantly from the training distribution.
    We first identify the most important numerical column, $j^{*}$, using a pre-trained XGBoost~\citep{XGBoost}. A Gaussian distribution
    \begin{equation*}
        \mathcal{N}(z | \mu_{j^{*}}^{s}, \sigma_{j^{*}}^{s}) = \frac{1}{\sqrt{2 \pi \sigma_{j^{*}}^{s}}} \exp\left(-\frac{(z - \mu_{j^{*}}^{s})^2}{2 (\sigma_{j^{*}}^{s})^2}\right)
    \end{equation*}
    is then fitted to the $j^{*}$-th column of the training data, using $\mu_{j^{*}}^{s}$ and $\sigma_{j^{*}}^{s}$.
    The likelihood of each test sample $\mathbf{x}_i^{t}$ is then computed as $\mathcal{N}(\mathbf{x}_{ij^{*}}^{t} | \mu_{j^{*}}^{s}, \sigma_{j^{*}}^{s})$.
    Finally, test samples are drawn inversely proportional to their likelihood, with the sampling probability $\mathbb{P}(\mathbf{x}_{i}^{t})$ of $\mathbf{x}_{i}^{t}$ is defined as:
    \begin{equation*}
        \mathbb{P}(\mathbf{x}_{i}^{t}) = \frac{\mathcal{N}(\mathbf{x}_{ij^{*}}^{t} | \mu_{j^{*}}^{s}, \sigma_{j^{*}}^{s})^{-1}} {\sum_{i'=1}^{|\mathcal{D}_t|} \mathcal{N}(\mathbf{x}_{i'j^{*}}^{t} | \mu_{j^{*}}^{s}, \sigma_{j^{*}}^{s})^{-1}}.
    \end{equation*}

    \item \textbf{Categorical:} Important categorical column shift simulates natural domain shifts where the test distribution of the most important categorical column deviates significantly from the training distribution.
    Again, we first identify the most important categorical column, $j^{*}$, using a pre-trained XGBoost~\citep{XGBoost}. A categorical distribution, which generalizes the Bernoulli distribution,
    \begin{equation*}
        \mathcal{C}(z | p_1, \cdots, p_K) = p_1^{\mathbbold{1}_{\{ 1 \}}(z)} \cdots p_K^{\mathbbold{1}_{\{ K \}}(z)},
    \end{equation*}
    is then fitted to the $j^{*}$-th column of the training data, where $K$ is the number of distinct categorical features in the $j^{*}$-th column, and $p_k = \mathbb{    P}(\bar{\mathbf{x}}_j^{s} = k)$ for $k = 1, \cdots, K$.
    The likelihood of each test sample $\mathbf{x}_i^{t}$ is then computed as $\mathcal{C}(\mathbf{x}_{ij^{*}}^{t} | p_1, \cdots, p_K)$.
    Finally, test samples are drawn inversely proportional to their likelihood, with the sampling probability $\mathbb{P}(\mathbf{x}_{i}^{t})$ of $\mathbf{x}_{i}^{t}$ is defined as:
    \begin{equation*}
        \mathbb{P}(\mathbf{x}_{i}^{t}) = \frac{\mathcal{C}(\mathbf{x}_{ij^{*}}^{t} | p_1, \cdots, p_K)^{-1}}{\sum_{i'=1}^{|\mathcal{D}_t|} \mathcal{C}(\mathbf{x}_{i'j^{*}}^{t} | p_1, \cdots, p_K)^{-1}}.
    \end{equation*}

\end{itemize}

\subsection{Label Distribution Shifts} \label{subsec:further_label_distribution_shifts}
\begin{itemize}
    \item \textbf{Class Imbalance:} This label distribution shift simulates a highly class-imbalanced test stream, where labels that are rare in the training set are more likely to appear frequently in the test set.
    Given a class imbalance ratio $\rho = 10$, we first rank the output labels $y_i^{t} \in \mathcal{Y}$ for each test sample $\mathbf{x}_{i}^{t}$ in ascending order of their frequency in the training set, assigning ranks from 1 to $C$, where $C$ is the number of classes. Specifically, $\text{rank}(y_{i}^{t}) = 1$ indicates that $y_i^{t}$ is the least frequent label in the training set, while $\text{rank}(y_{i}^{t}) = C$ indicates that $y_i^{t}$ is the most frequent.
    We then define the unnormalized sampling probability for each test sample $\mathbf{x}_{i}^{t}$ as:
    \begin{equation*}
        \tilde{\mathbb{P}}(\mathbf{x}_{i}^{t}) = \frac{\text{rank}(y_i^{t})}{C} (\rho - 1) + 1.
    \end{equation*}
    The normalized sampling probability $\mathbb{P}(\mathbf{x}_{i}^{t})$ for each test sample $\mathbf{x}_{i}^{t}$ is then defined as:
    \begin{equation*}
        \mathbb{P}(\mathbf{x}_{i}^{t}) = \frac{\tilde{\mathbb{P}}(\mathbf{x}_{i}^{t})}{\sum_{i' = 1}^{|\mathcal{D}_t|} \tilde{\mathbb{P}}(\mathbf{x}_{i'}^{t})}.
    \end{equation*}

    \item \textbf{Temporal Correlation:} To simulate temporal correlations in test data, we employ a custom sampling strategy using the Dirichlet distribution. This approach effectively captures temporal dependencies by dynamically adjusting the label distribution over time. We begin with a uniform probability distribution $\mathbb{P}_0 = \left( 1 / C \right)_{j=1}^{C}$, where $C$ is the number of classes. For sampling the $i$-th test instance, a probability distribution $\boldsymbol{\pi}_i$ is drawn from the Dirichlet distribution:
    \begin{equation*}
        \boldsymbol{\pi}_i \sim \text{Dirichlet}(\mathbb{P}_{i-1}),
    \end{equation*}
    and then smoothed using $\eta = 10^{-6}$ to avoid zero probabilities for any class $j$:
    \begin{equation*}
        \boldsymbol{\pi}_i = \frac{\max(\eta, \boldsymbol{\pi}_i)}{\sum_{j=1}^{C} \max(\eta, \boldsymbol{\pi}_{ij})}.
    \end{equation*}
    A label $y_i^{t}$ is subsequently sampled according to $\boldsymbol{\pi}_i$, and the corresponding test instance $\mathbf{x}_{i}^{t}$ is randomly selected from the test data with label $y_i^{t}$. After the $i$-th sampling, the distribution $\mathbb{P}_i$ is updated using the recent history of sampled labels within a sliding window of size $w = 5$:
    \begin{equation*}
        \mathbb{P}_i \leftarrow \bigg( \frac{1}{w} \sum_{i'=i-w+1}^{i} \mathbbold{1}_{\{ j \}}(y_{i'}^{t}) \bigg)_{j=1}^{C}.
    \end{equation*}
    
\end{itemize}
%%%%%%%%%%%%%%%%%%%%%%%%%%%%%%%%%%%%%%%%%%%%%%%%%%%%%%%%%%%%

%%%%%%%%%%%%%%%%%%%%%%%%%%%%%%%%%%%%%%%%%%%%%%%%%%%%%%%%%%%%
\section{Baseline Details} \label{sec:baselines}

\subsection{Deep Tabular Learning Architectures}

\begin{itemize}
    \item \textbf{MLP:} Multi-Layer Perceptron (MLP)~\citep{mlp} is a foundational deep learning architecture characterized by multiple layers of interconnected nodes, where each node applies a non-linear activation function to a weighted sum of its inputs. In the tabular domain, MLP is often employed as a default deep learning model, with each input feature corresponding to a node in the input layer.

    \item \textbf{AutoInt:} Automatic Feature Interaction Learning via Self-Attentive Neural Networks (AutoInt)~\citep{song2019autoint} is a model that automatically learns complex feature interactions in tasks like click-through rate (CTR) prediction, where features are typically sparse and high-dimensional. It uses a multi-head self-attentive neural network to map features into a low-dimensional space and capture high-order combinations, eliminating the need for manual feature engineering. AutoInt efficiently handles large datasets, outperforms existing methods, and provides good explainability.
    
    \item \textbf{ResNet:} ResNet for tabular data~\citep{fttransformer}, is a modified version of the original ResNet architecture~\citep{he2016deep}, tailored to capture intricate patterns within structured datasets. Although earlier efforts yielded modest results, recent studies have re-explored ResNet's capabilities, inspired by its success in computer vision and NLP. This ResNet-like model for tabular data is characterized by a streamlined design that facilitates optimization through nearly direct paths from input to output, enabling the effective learning of deeper feature representations.

    \item \textbf{FT-Transformer:} Feature Tokenizer along with Transformer (FT-Transformer)~\citep{fttransformer}, represents a straightforward modification of the Transformer architecture tailored for tabular data. In this model, the feature tokenizer component plays a crucial role by converting all features, whether categorical or numerical, into tokens. Subsequently, a series of Transformer layers are applied to these tokens within the Transformer component, along with the added {[CLS]} token. The ultimate representation of the {[CLS]} token in the final Transformer layer is then utilized for the prediction.

\end{itemize}

\subsection{Supervised Baselines} \label{subsec:sup_baselines}

\begin{itemize}
    \item \textbf{$k$-NN:} $k$-Nearest Neighbors ($k$-NN) is a fundamental model in tabular learning that identifies the $k$ closest data points based on a chosen metric. It makes predictions through majority voting for classification or weighted averaging for regression. The hyperparameter $k$ influences the model's sensitivity.

    \item \textbf{LogReg:} Logistic Regression (LogReg) is a linear classification model that estimates the probability of class membership using a logistic function, which maps the linear combination of features to a range of $[0, 1]$. With proper regularization, LogReg can achieve performance comparable to state-of-the-art tabular models.
    
    \item \textbf{RandomForest:} Random Forest is an ensemble learning algorithm that builds multiple decision trees to improve accuracy and reduce overfitting. It is particularly effective at capturing non-linear patterns and is robust against outliers.
    
    \item \textbf{XGBoost:} Extreme Gradient Boosting (XGBoost)~\citep{XGBoost} is a boosting algorithm that sequentially builds weak learners, typically decision trees, to correct errors made by previous models. XGBoost is known for its high predictive performance and ability to handle complex relationships through regularization.
    
    \item \textbf{CatBoost:} CatBoost~\citep{catboost}, like XGBoost, is a boosting algorithm that excels in handling categorical features without extensive preprocessing. It is highly effective in real-world datasets, offering strong performance, albeit at the cost of increased computational resources and the need for parameter tuning.

\end{itemize}

\subsection{Test-Time Adaptation Baselines} \label{subsec:tta_baselines}

\begin{itemize}
    \item \textbf{PL:} Pseudo-Labeling (PL)~\citep{pl} leverages a pseudo-labeling strategy to update model parameters during test time.
    
    \item \textbf{TTT++:} Improved Test-Time Training (TTT++)~\citep{liu2021ttt++} enhances test-time adaptation by using feature alignment strategies and regularization, eliminating the need to access source data during adaptation.
    
    \item \textbf{TENT:} Test ENTropy minimization (TENT)~\citep{wang2020tent} updates the scale and bias parameters in the batch normalization layer during test time by minimizing entropy within a given test batch.
    
    \item \textbf{EATA:} Efficient Anti-forgetting Test-time Adaptation (EATA)~\citep{niu2022efficient} mitigates the risk of unreliable gradients by filtering out high-entropy samples and applying a Fisher regularizer to constrain key model parameters during adaptation.
    
    \item \textbf{SAR:} Sharpness-Aware and Reliable optimization (SAR)~\citep{niu2023towards} builds on TENT by filtering samples with large entropy, which can cause model collapse during test time, using a predefined threshold.
    
    \item \textbf{LAME:} Laplacian Adjusted Maximum-likelihood Estimation (LAME)~\citep{lame} employs an output adaptation strategy during test-time, focusing on adjusting the model's output probabilities rather than tuning its parameters.

\end{itemize}
%%%%%%%%%%%%%%%%%%%%%%%%%%%%%%%%%%%%%%%%%%%%%%%%%%%%%%%%%%%%

%%%%%%%%%%%%%%%%%%%%%%%%%%%%%%%%%%%%%%%%%%%%%%%%%%%%%%%%%%%%
\section{Further Experimental Details} \label{sec:exp_details}

\subsection{Further Implementation Details}

All experiments are conducted on two servers. The first server is equipped with a 40-core Intel Xeon E5-2630 v4 CPU, 252GB RAM, 4 NVIDIA TITAN Xp GPUs, and runs Ubuntu 18.04.4. The second server has a 40-core Intel Xeon E5-2640 v4 CPU, 128GB RAM, 8 NVIDIA TITAN Xp GPUs, and runs Ubuntu 22.04.4. All architectures were implemented using Python 3.8.16 with PyTorch~\citep{paszke2019pytorch} and PyTorch Geometric~\citep{fey2019fast}. % The specific versions of all software libraries and frameworks used are provided in the \texttt{AdapTable/requirements.txt} file of the supplementary materials. We also include our source code in \texttt{AdapTable} folder of the supplementary materials. Please refer to this for all experimental details and to clarify any uncertainties.

% \ch{including GPU/CPU models; amount of memory; operating system; names and versions of relevant software libraries and frameworks. (yes)} \\
% \ch{(이게 꼭 필요하구나) Analysis of experiments goes beyond single-dimensional summaries of performance (e.g., average; median) to include measures of variation, confidence, or other distributional information.} \\
% \ch{The significance of any improvement or decrease in performance is judged using appropriate statistical tests (e.g., Wilcoxon signed-rank).}

\subsection{Hyperparameters for Supervised Baselines}

For $k$-NN, LogReg, RandomForest, XGBoost, and CatBoost, optimal parameters are determined for each dataset using a random search with 10 iterations on the validation set. The search space for each method is specified in Table~\ref{table:sup_hyperparam_searchspace}.

\input{tables/hparam_space_sup}

\subsection{Hyperparameters for TTA Baselines} \label{subsec:tta_hparams}

In scenarios where the test set is unknown, tuning the hyperparameters of TTA methods on the test set would be considered cheating. Therefore, we tune all hyperparameters for each TTA method and backbone classifier architecture using the Numerical common corruption on the CMC tabular dataset, which we did not use as test data in OpenML-CC18~\citep{openml_benchmark} benchmark.
PL, TENT~\citep{wang2020tent}, and SAR~\citep{niu2023towards} require three main hyperparameters---learning rate, number of adaptation steps per batch, and the option for episodic adaptation, where the model is reset after each batch. PL~\citep{pl} and TENT use a learning rate of 0.0001 with 1 adaptation step and episodic updates. Additionally, SAR requires a threshold to filter high-entropy samples and is configured with a learning rate of 0.001, 1 adaptation step, and episodic updates.
For TTT++~\citep{liu2021ttt++}, EATA~\citep{niu2022efficient}, and LAME~\citep{lame}, we follow the authors' hyperparameter settings, except for the learning rate and adaptation steps. TTT++ and EATA were configured with a learning rate of 0.00001, 10 adaptation steps, and episodic updates. LAME, which only adjusts output logits, does not require hyperparameters related to gradient updates.
For all baselines, hyperparameter choices remained consistent across different architectures, including MLP, AutoInt, ResNet, and FT-Transformer. The hyperparameter search space % and selected values
for each method are detailed in Table~\ref{table:tta_hyperparam_searchspace}. % and Table~\ref{table:tta_selected_hyperparam}, respectively.

\input{tables/hparam_space_tta}

\subsection{Hyperparameters for AdapTable}

AdapTable requires three test-time hyperparameters---the smoothing factor $\alpha$, and the low and high uncertainty quantiles $q_{\text{low}}$ and $q_{\text{high}}$. % The specific parameters for each backbone classifier architecture are detailed in Table~\ref{table:adaptable_hyperparameter_selection}.
For fairness, we tune all AdapTable hyperparameters across different backbone architectures using the Numerical common corruption on the CMC dataset from the OpenML-CC18 benchmark~\citep{openml_benchmark}, which is not used as test data. We observe that AdapTable's hyperparameter choices remain consistent across various architectures, including MLP, AutoInt, ResNet, and FT-Transformer. Notably, AdapTable demonstrates high insensitivity to variations in $\alpha$, $q_{\text{low}}$, and $q_{\text{high}}$, which are uniformly set to 0.1, 0.25, and 0.75, respectively, across all datasets and architectures.

% AdapTable requires three test-time hyperparameters: the smoothing factor $\alpha$, and the low and high uncertainty quantiles $q_{\text{low}}$ and $q_{\text{high}}$. The parameters specific to each backbone classifier architecture are outlined in Table~\ref{table:adaptable_hyperparameter_selection}.

% For fairness and integrity, we also tune all hyperparameters for AdapTable across different backbone classifier architectures using the Numerical common corruption on the CMC tabular dataset, which we did not use as test data in OpenML-CC18~\citep{openml_benchmark} benchmark. We also find that hyperparameter choices of AdapTable remains consistent across different architectures, including MLP, AutoInt, ResNet, and FT-Transformer.

% It is worth noting that AdapTable is highly insensitive to variations in all of these hyperparameters $\alpha$, $q_{\text{low}}$, and $q_{\text{high}}$, where we set them as 0.1, 0.25, and 0.75 across different datasets and classifier architectures, respectively.

% \input{tables/hparam_selected_ours}
%%%%%%%%%%%%%%%%%%%%%%%%%%%%%%%%%%%%%%%%%%%%%%%%%%%%%%%%%%%%

%%%%%%%%%%%%%%%%%%%%%%%%%%%%%%%%%%%%%%%%%%%%%%%%%%%%%%%%%%%%
\section{Additional Analysis} \label{sec:further_analysis}

\subsection{Latent Space Visualizations}

In Figure~\ref{fig:further_latent}, we further visualize latent spaces of test instances using t-SNE across six different datasets and four representative deep tabular learning architectures to illustrate the observation discussed in Section~\ref{subsec:em_failure}. This visualization highlights the complex decision boundaries within the latent space of tabular data, which are significantly more intricate than those observed in other domains. By comparing the upper four rows---HELOC, Voting, Hospital Readmission, and Childhood Lead---with the lower two rows---linearized image data (MFEAT-PIXEL) and homogeneous DNA string sequences (DNA)---it becomes evident that the latent space decision boundaries in the tabular domain are particularly complex. According to WhyShift~\cite{whyshift}, this complexity is primarily due to latent confounders inherent in tabular data and concept shifts, where such confounders cause output labels to vary greatly for nearly identical inputs. As discussed in Section~\ref{subsec:em_failure}, this further underscores the limitations of existing TTA methods~\citep{sun2020test,gandelsman2022test,liu2021ttt++,lame,ods}, which often depend on the cluster assumption.

\subsection{Reliability Diagrams}

Figure~\ref{fig:further_reliability} presents additional reliability diagrams across five different datasets and four representative deep tabular learning architectures, illustrating that tabular data often displays a mix of overconfident and underconfident prediction patterns. This contrasts with the consistent overconfidence observed in the image domain~\citep{isotonic} and underconfidence in the graph domain~\citep{wang2021confident}. As shown in Figure~\ref{fig:further_reliability}, the Voting and Hospital Readmission datasets consistently exhibit overconfident behavior across all architectures, while the HELOC, Childhood Lead, and Diabetes datasets demonstrate underconfident tendencies. These observations underscore the need for a tabular-specific uncertainty calibration method.

\subsection{Label Distribution Shifts and Prediction Bias Towards Source Label Distributions}

We demonstrate that the data distribution shift we primarily target in the tabular domain---label distribution shift---occurs frequently in practice. Figure~\ref{fig:further_label_distribution} presents the source label distribution (a), target label distribution (b), pseudo label distribution for test data using the source model (c), and the estimated target label distribution after applying our label distribution handler (d) across the five datasets. Comparing (a) and (b) in each row, it is evident that label distribution shift occurs across all datasets. In (c), we observe that the marginal label distribution predicted by the source model is commonly biased towards the source label distribution. Lastly, (d) illustrates that our label distribution handler effectively estimates the target label distribution, guiding the pseudo label distribution towards the target label distribution.

\subsection{Entropy Distributions}

We highlight a unique characteristic of tabular data: model prediction entropy consistently shows a strong bias toward underconfidence. To illustrate this, we present entropy distribution histograms for test instances across six datasets and four representative deep tabular learning architectures in Figure~\ref{fig:further_entropy}. A clear pattern emerges when comparing the upper four rows (HELOC, Voting, Hospital Readmission, Childhood Lead) with the lower two (Optdigits, DNA). The upper rows exhibit consistently high entropy, indicating a skew toward underconfidence, while the lower rows do not, except for Childhood Lead, where extreme class imbalance causes the model to collapse to the major class. This analysis highlights the distinct bias of tabular data toward underconfident predictions, a pattern less common in other domains. This aligns with findings that applying unsupervised objectives like entropy minimization to high-entropy samples can result in gradient explosions and model collapse~\citep{niu2023towards}.
%%%%%%%%%%%%%%%%%%%%%%%%%%%%%%%%%%%%%%%%%%%%%%%%%%%%%%%%%%%%

%%%%%%%%%%%%%%%%%%%%%%%%%%%%%%%%%%%%%%%%%%%%%%%%%%%%%%%%%%%%
\section{Additional Experiments} \label{sec:further_experiments}

\input{figures/various_architectures}

\subsection{Result Across Diverse Model Architectures}
In Figure~\ref{fig:various_architectures}, we report AdapTable's effectiveness across three mainstream tabular learning architectures---AutoInt~\citep{song2019autoint}, ResNet~\citep{fttransformer}, and FT-Transformer~\citep{fttransformer}. We report the average macro F1 score across three datasets---HELOC, Voting, and Childhood Lead. None of the baselines outperform the original source model, with LAME~\citep{lame} even showing significant performance drops. In contrast, AdapTable consistently achieves significant improvements across all architectures, highlighting its robustness and versatility.

\subsection{Further Analysis} \label{exp:analysis}

\paragraph{Computational efficiency.}
The leftmost part of Figure~\ref{fig:efficieny_sensitivity} compares the computational efficiency of AdapTable with TTA baselines. On the HELOC dataset, AdapTable's total elapsed time is approximately 1.54 seconds, translating to about 0.0002 seconds per sample, which is highly desirable. Moreover, AdapTable achieves an optimal efficiency-efficacy trade-off.

\paragraph{Hyperparameter sensitivity.}
Figure~\ref{fig:efficieny_sensitivity} further analyzes the hyperparameter sensitivity of AdapTable on the Childhood Lead dataset. As shown in the figure, AdapTable remains highly insensitive to changes in the smoothing factor $\alpha$, low uncertainty quantile $q_{\text{low}}$, and high uncertainty quantile $q_{\text{high}}$.

\input{figures/efficiency_sensitivity}

\subsection{Detailed Results Across Common Corruptions and Datasets}

Figure~\ref{fig:main_common_corruption} presents the average F1 score across six types of common corruption and three datasets. Here, we provide more detailed results, including the standard errors. As shown in Table~\ref{table:further_common_corruptions}, AdapTable outperforms baseline TTA methods by a large margin across all datasets and corruption types. This further highlights the empirical efficacy of AdapTable, not only in handling label distribution shifts but also in addressing various common corruptions.

\subsection{All Results Across Datasets and Model Architectures}

In Figure~\ref{fig:various_architectures}, we demonstrate the effectiveness of AdapTable across various tabular model architectures by reporting the average performance across three datasets. Here, we provide the mean and standard error for each dataset and architecture. As shown in Table~\ref{table:further_architecture}, AdapTable consistently achieves state-of-the-art performance with significant improvements across all model architectures and datasets. This further underscores the versatility and robustness of AdapTable.

\subsection{Additional Computational Efficiency Analysis}

One may wonder whether the post-training time required for AdapTable's shift-aware uncertainty calibrator is prohibitively long. To address this concern, we measure and report the elapsed real time for post-training our shift-aware uncertainty calibrator on the medium-scale Hospital Readmission dataset using the FT-Transformer architecture. The post-training process takes approximately 9.2 seconds. For small- and medium-scale datasets, the post-training process typically requires only a few seconds, and even in our largest experimental setting, the time remains minimal, taking at most a few minutes.
%%%%%%%%%%%%%%%%%%%%%%%%%%%%%%%%%%%%%%%%%%%%%%%%%%%%%%%%%%%%

%%%%%%%%%%%%%%%%%%%%%%%%%%%%%%%%%%%%%%%%%%%%%%%%%%%%%%%%%%%%
\section{Limitations and Broader Impacts}

\subsection{Limitations}

Similar to other test-time training (TTT) methods~\cite{sun2020test,liu2021ttt++,gandelsman2022test}, AdapTable requires an additional post-training stage to integrate a shift-aware uncertainty calibrator during the source model's training phase. While full test-time adaptation methods~\cite{wang2020tent,niu2022efficient,niu2023towards} avoid this, our analysis in Section~\ref{subsec:em_failure} and experiments in Section~\ref{sec:exp} show that they fail in the tabular domain due to their focus on input covariate shifts, which are often entangled with concept shifts. According to WhyShift~\cite{whyshift}, concept shifts, driven by changes in latent confounders, require natural language descriptions of the shift conditions, necessitating a data-centric approach. Additionally, while AdapTable performs well across various corruptions beyond label distribution shifts (Figure~\ref{fig:main_common_corruption}), it is primarily focused on addressing label distribution shifts. Further exploration is needed to assess its effectiveness in handling input covariate shifts or concept shifts.

\subsection{Broader Impacts}

Tabular data is prevalent across industries such as healthcare~\citep{johnson2016mimic,mimiciv_v1}, finance~\citep{FICO,anes}, manufacturing~\citep{manufacturing}, and public administration~\citep{tableshift}. Our research addresses the critical yet underexplored challenge of distribution shifts in tabular data, a problem that has not received sufficient attention. We believe that our approach can significantly enhance the performance of machine learning models in various industries by improving model adaptation to tabular data, thereby creating meaningful value in practical applications.
Through our data-centric analysis in Section~\ref{sec:analysis}, we identify why existing TTA methods fail in the tabular domain and introduce a tabular-specific approach for handling label distribution shifts in Section~\ref{sec:method}. We hope this work will provide valuable insights for future research on test-time adaptation in tabular data. Additionally, by making our source code publicly available, we aim to support real-world applications across various fields, benefiting both academia and industry.

\input{tables/further_common_corruption}
\input{tables/further_architecture}
\input{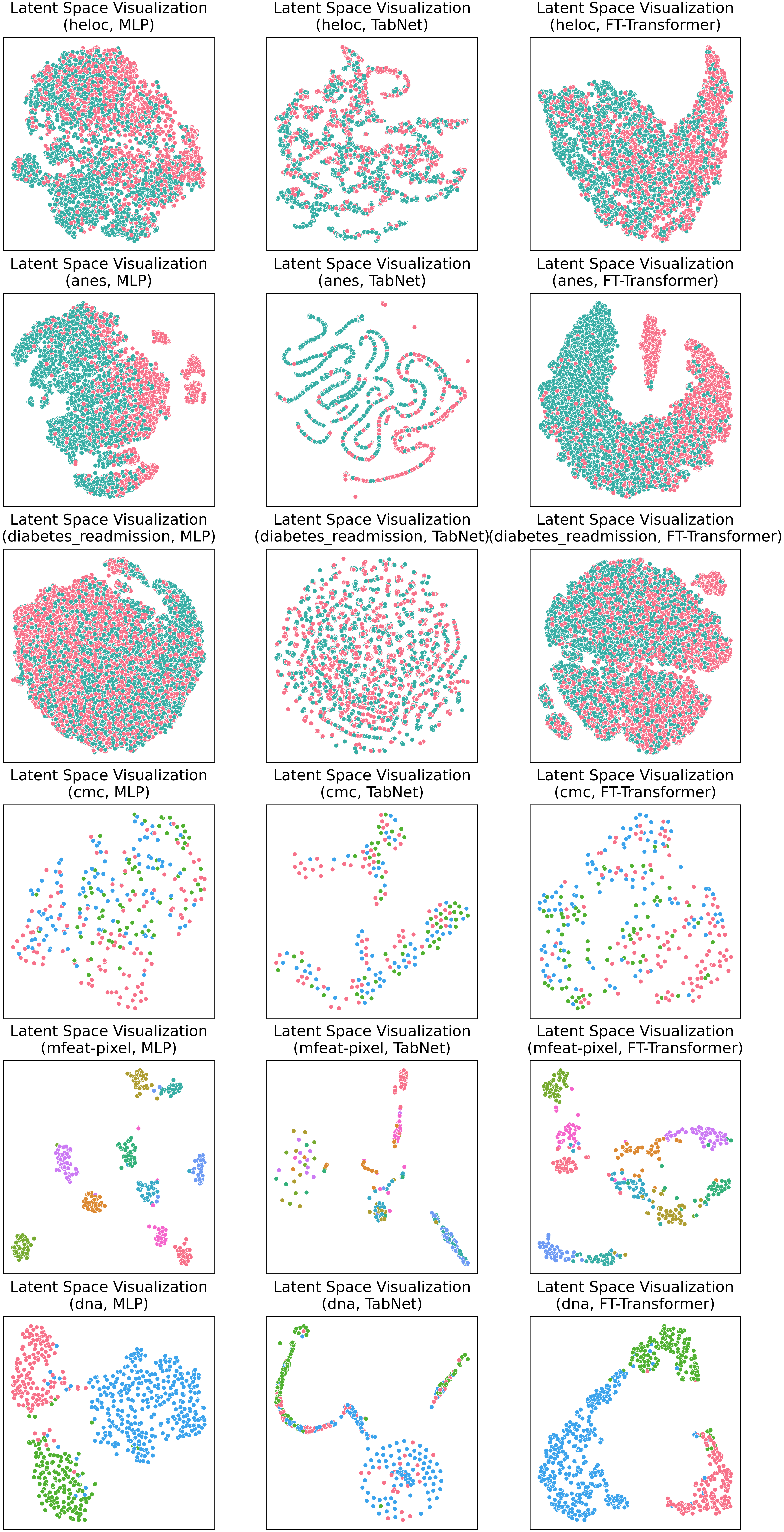}
\input{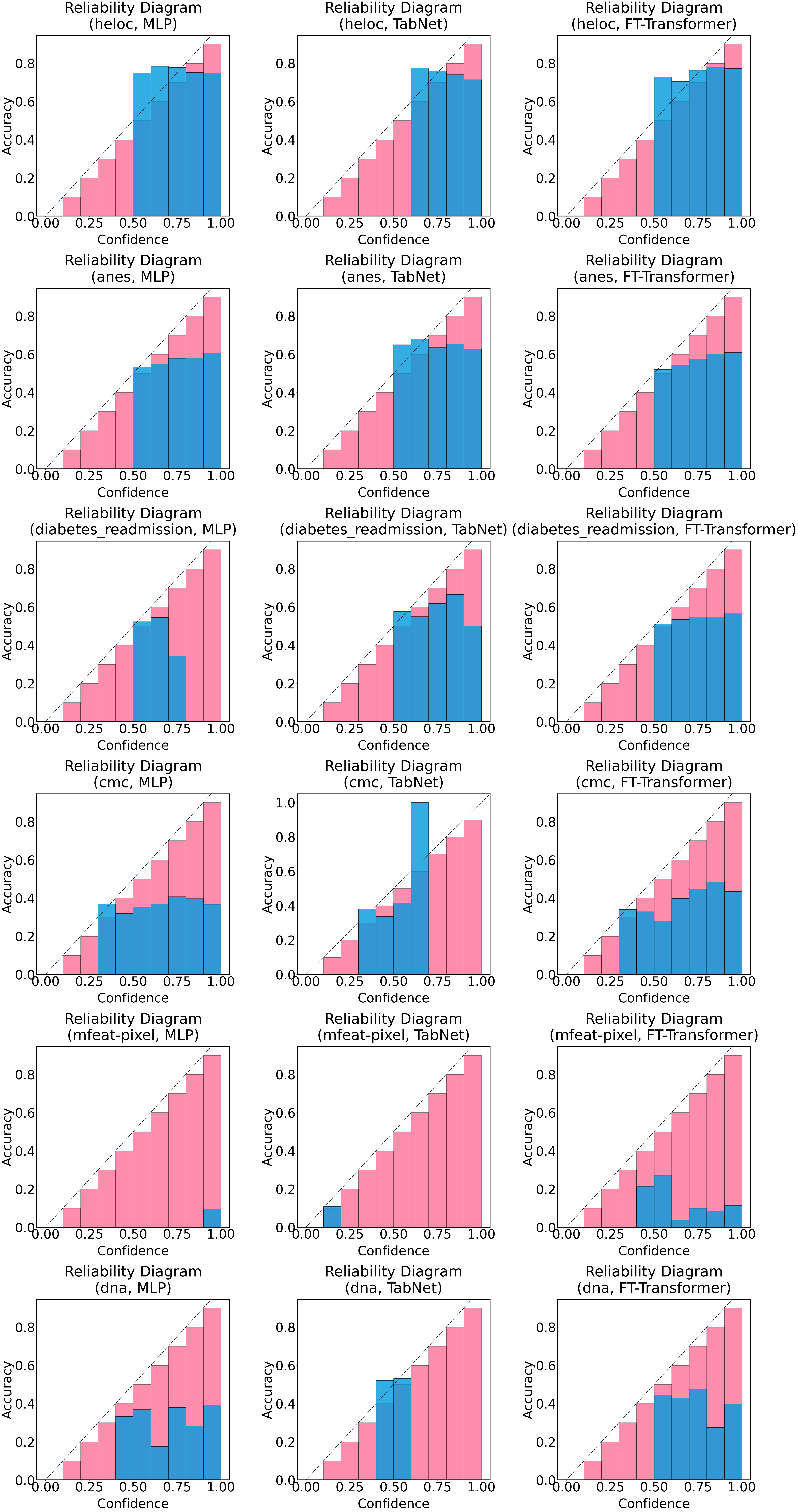}
\input{figures/further_label_distribution}
\input{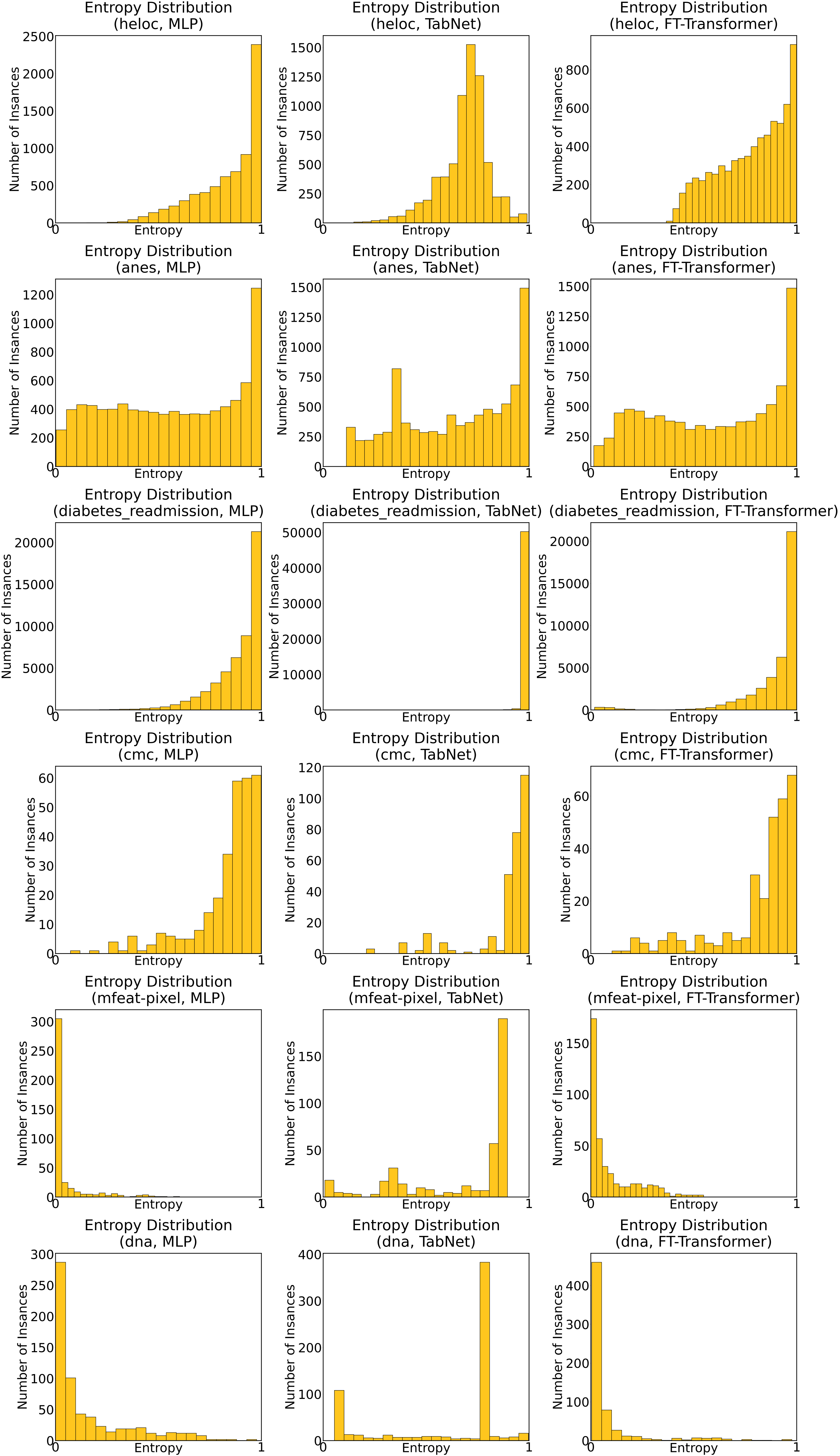}

%% file: sections/5_related_work.tex
\section{Related Work}

\paragraph{Machine learning for tabular data.}
The distinct nature of tabular data reduces the effectiveness of deep neural networks, making gradient-boosted decision trees~\citep{XGBoost,catboost} more suitable. However, research continues to develop deep learning models tailored for tabular data~\citep{mlp,song2019autoint,arik2021tabnet,fttransformer}, including recent efforts involving large language models~\citep{fang2024large,hegselmann2023tabllm,hollmann2023tabpfn,dinh2022lift} that leverage textual prior knowledge. Notably, our method is architecture-agnostic and can be applied to any model.

\paragraph{Distribution shifts in the tabular domain.}
Recently, distribution shift benchmarks for tabular data have been introduced~\citep{whyshift,tableshift}. WhyShift~\citep{whyshift} reveals that concept shifts ($Y|X$-shifts) are more prevalent and detrimental than covariate shifts ($X$-shifts). TableShift~\citep{tableshift} offers a benchmark with 15 classification tasks, highlighting a strong correlation between shift gaps and label distribution shifts ($Y$-shifts), which supports the validity of our method.

\paragraph{Test-time adaptation.}
Over the past years, test-time adaptation (TTA) methods have been proposed across various domains, such as computer vision~\citep{wang2020tent,gong2022note,niu2023towards,cloudfixer}, natural language processing~\citep{shi2024medadapter,liang2024comprehensive}, and speech processing~\citep{sgem}. These methods adapt pre-trained models to unlabeled target domains without requiring access to source data, making them well-suited for sensitive tabular data. TabLog~\citep{tablog} is a recent TTA method specifically for tabular data, but it has architectural constraints and lacks a comprehensive analysis of distribution shifts. This underscores the need for model-agnostic TTA methods with a deeper understanding of tabular data, which we address in this paper.

%% file: algorithms/adaptable.tex
\begin{algorithm*}[!t]
\caption{AdapTable} \label{alg:adaptable}
\begin{algorithmic}[1]

\State \textbf{Input:} Pre-trained classifier $f_{\theta}(\cdot)$, post-trained shift-aware uncertainty calibrator $g_{\phi}(\cdot, \cdot)$, indicator function $\mathbbold{1}_{(\cdot)}(\cdot)$, quantile function $Q(\cdot, \cdot)$, softmax function $\text{softmax}(\cdot)$, normalization function $\text{norm}(\cdot)$, source data $\mathcal{D}_s = {\{( \mathbf{x}_i^s, y_i^s )\}}_{i}$, current test batch ${\{ \mathbf{x}_{i}^{t} \}}_{i=1}^{N}$, source class imbalance ratio $\rho = \max_{j} p_{s}(y)_{j} / \min_{j} p_{s}(y)_{j}$

\State \textbf{Parameters:} Smoothing factor $\alpha$, low uncertainty quantile $q_{\text{low}}$, high uncertainty quantile $q_{\text{high}}$

\State $p_s(y),~T \leftarrow {\big( \frac{1}{|\mathcal{D}_{s}|} \sum_{i=1}^{|\mathcal{D}_s|}{\mathbbold{1}_{ \{ j \} }(y_i^s)} \big)}_{j=1}^{C},~1.5\rho / (\rho - 1 + 10^{-6})$

\For{$u = 1$ to $D$}
    \State $\mathbf{s}_{u}^{t} \leftarrow \big( \mathbf{x}^{t}_{iu} - \frac{1}{|\mathcal{D}_{s}|} \sum_{i'=1}^{|\mathcal{D}_s|}{\mathbf{x}_{i'u}^s} \big)_{i=1}^{N}$
    \Comment{Compute shift trend $\textbf{s}^{t}$}
\EndFor

\For{$i = 1$ to $N$}
    \State $p_t(y|\mathbf{x}_{i}^{t}) \leftarrow \text{softmax}\big(f_{\theta}(\mathbf{x}_i^{t}) \big)$

    \State $T_i \leftarrow g_{\phi}\big( f_{\theta}(\mathbf{x}_i^t), \mathbf{s}^t \big)$ 
    \Comment{Determine per-sample temperature $\mathbf{x}_i^t$}

    \State $j^{*},~j^{**} \leftarrow \argmax_{1 \le j \le C} {p_t(y|\mathbf{x}_{i}^{t})}_{j},~\argmax_{1 \le j \le C, j \ne j^{*}} {p_t(y|\mathbf{x}_{i}^{t})}_{j}$
    
    \State $\delta_i \leftarrow {\big( { \text{softmax}\big(f_{\theta}(\mathbf{x}_i^{t}) / T_i\big) }_{j^{*}} - { \text{softmax}\big(f_{\theta}(\mathbf{x}_i^{t}) / T_i \big) }_{j^{**}} \big)}^{-1}$
    \Comment{Define uncertainty of $\mathbf{x}_i^t$ as a margin of $f_{\theta}(\mathbf{x}_i^{t}) / T_i$}

    \State $p^{\text{de}}_{t}(y|\mathbf{x}_{i}^{t}) \leftarrow \text{norm}\big(p_t(y|\mathbf{x}_{i}^{t}) / p_s(y)\big)$
    \Comment{Compute debiased target label estimator}
\EndFor

\State $p_t(y) \leftarrow (1 - \alpha) \cdot \frac{1}{N} {\sum_{i=1}^{N} p^{\text{de}}_{t}(y|\mathbf{x}_{i}^{t})} + \alpha \cdot {p^{\text{oe}}_t(y)}$
\Comment{Estimate target label distribution}

\For{$i = 1$ to $N$}
    \If{$\delta_{i} \ge Q\big(\{\delta_{i'}\}_{i'=1}^{N}, q_{\text{high}}\big)$}
        \State $\tilde{T}_i \leftarrow T$
    \ElsIf{$\delta_{i} \le Q\big(\{\delta_{i'}\}_{i'=1}^{N}, q_{\text{low}}\big)$}
        \State $\tilde{T}_i \leftarrow 1/T$
        \Comment{Calculate temperature $\tilde{T}_i$ using uncertainty $\delta_i$}
    \Else
        \State $\tilde{T}_i \leftarrow 1$
    \EndIf

    \State $\tilde{p}_t(y|\mathbf{x}_{i}^{t}) \leftarrow \text{softmax}\big( f_{\theta}(\mathbf{x}_i^t) / \tilde{T}_i \big)$
    \Comment{Perform temperature scaling with $\tilde{T}_i$}
    
    \State $\bar{p}_{t}(y|\mathbf{x}_{i}^{t}) \leftarrow \big( \tilde{p}_t(y|\mathbf{x}_{i}^{t}) + \text{norm}( \tilde{p}_t(y|\mathbf{x}_{i}^{t}) p_t(y) / p_s(y)) \big) / 2$
    \Comment{Perform self-ensembling}
\EndFor

\State $p^{\text{oe}}_{t}(y) \leftarrow (1 - \alpha) \cdot \frac{1}{N} \sum_{i=1}^{N}{\bar{p}_{t}(y|\mathbf{x}_{i}^{t})} + \alpha \cdot p^{\text{oe}}_{t}(y)$
\Comment{Update online target label estimator}

\State \textbf{Output:} Final predictions ${\{\bar{p}_{i}(y)\}}_{i=1}^{N}$

\end{algorithmic}
\end{algorithm*}

%% file: tables/dataset_details.tex
\begin{table}[!h]
\centering
\caption{
    Summary of the datasets used in our experiments, including the total number of instances ({Total Samples}), the number of instances allocated to training, validation, and test sets ({Training Samples}, {Validation Samples}, {Test Samples}), the total number of features ({Total Features}), and a breakdown into numerical and categorical features ({Numerical Features}, {Categorical Features}). All tasks involve binary classification.
}
\label{table:dataset_specification}
\begin{adjustbox}{width=\linewidth}
\begin{tabular}{lrrrrrr}
    \toprule
    
    Statistic & HELOC & Voting & Hospital Readmission & ICU Mortality & Childhood Lead & Diabetes \\ \midrule
    
    Total Samples & 9,412 & 60,376 & 89,542 & 21,549 & 24,749 & 1,299,758 \\

    Training Samples & 2,220 & 34,796 & 34,288 & 7,116 & 11,807 & 969,229 \\
    
    Validation Samples & 278 & 4,349 & 4,286 & 889 & 1,476 & 121,154 \\
    
    Test Samples & 6,914 & 21,231 & 50,968 & 13,544 & 11,466 & 209,375 \\ \midrule

    Total Features & 22 & 54 & 46 & 7491 & 7 & 25 \\
    
    Numerical Features & 20 & 8 & 12 & 7490 & 4 & 6 \\
    
    Categorical Features & 2 & 46 & 34 & 1 & 3 & 19 \\
    
    \bottomrule
\end{tabular}
\end{adjustbox}
\end{table}

%% file: tables/hparam_space_sup.tex
\begin{table*}[!ht]
\centering
\caption{
    Hyperparameter search space of supervised baselines. \# neighbors denotes the number of neighbors, \# estim denotes the number of estimators, depth denotes the maximum depth, and lr denotes the learning rate, respectively.
}
\label{table:sup_hyperparam_searchspace}
\vspace{-.05in}
\begin{adjustbox}{width=\linewidth}
\begin{tabular}{ll}
    \toprule
    
    Method & Search Space \\ \midrule
    
    $k$-NN & \# neighbors: $\{2, \cdots, 12\}$ \\
    
    RandomForest & \# estim: $\{50, 100, 150, 200\}$, depth: $\{2, 3, \cdots, 12\}$ \\
    
    XGBoost & \# estim: $\{50, 100, 150, 200\}$, depth: $\{2, 3, \cdots, 12\}$, lr: $\{0.01, 0.01 + (1 - 0.01) / 19, \cdots, 1\}$, gamma: $\{0, 0.05, \cdots, 0.5\}$ \\
    
    CatBoost & \# iterations: $\{50, 100, \cdots, 2000\}$, lr: $\{0.01, 0.01 + (1 - 0.01) / 19, \cdots, 1\}$, depth: $\{5, \cdots, 40\}$ \\

    \bottomrule
\end{tabular}
\end{adjustbox}
\end{table*}

%% file: tables/hparam_space_tta.tex
\begin{table}[!ht]
\centering
\caption{
    Hyperparameter search space of test-time adaptation baselines. Here, we only denote the common hyperparameters, where method-specific hyperparameters are specified in Section~\ref{subsec:tta_hparams}.
}
\label{table:tta_hyperparam_searchspace}
\vspace{.05in}
\begin{tabular}{ll}
    \toprule
    
    Hyperparameter &  Search Space \\ \midrule
    
    lr & $\{10^{-3}, 10^{-4}, 10^{-5}, 10^{-6}\}$ \\
    
    \# steps & $\{1, 5, 10, 15, 20\}$ \\
    
    episodic & \{True, False\} \\
    
    \bottomrule
\end{tabular}
\end{table}

%% file: figures/various_architectures.tex
\begin{figure}[!t]
\centering
\includegraphics[width=.8\linewidth]{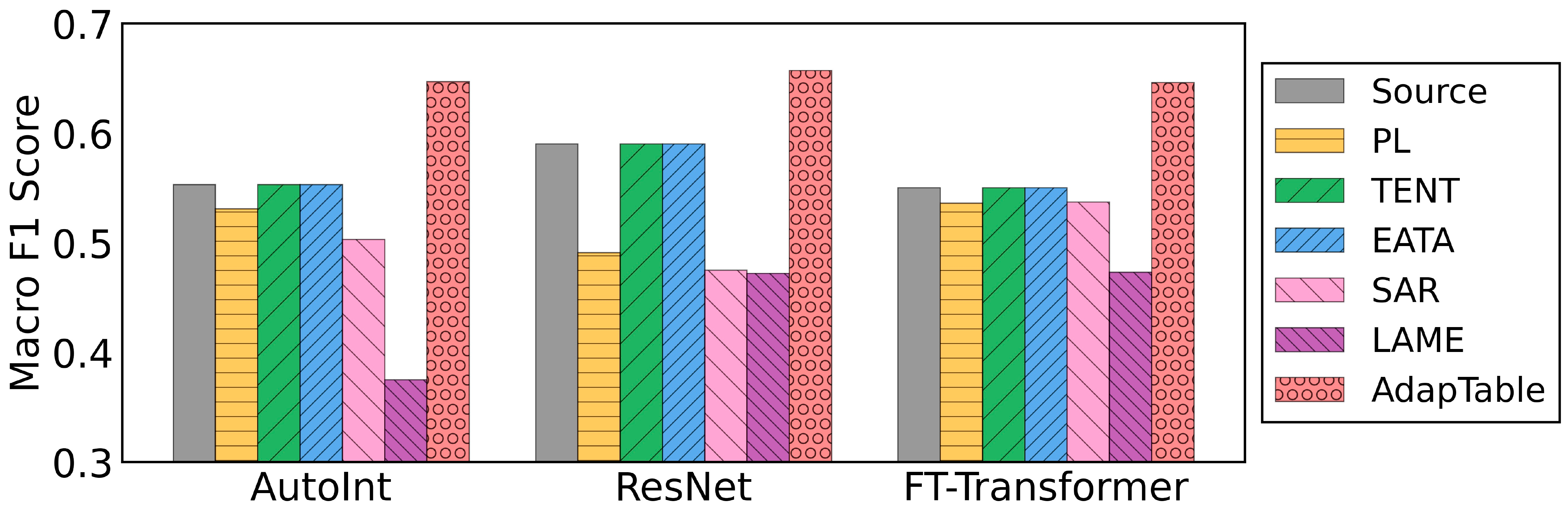}
\caption{
    The average macro F1 score of AdapTable and TTA baselines across three datasets (HELOC, Voting, Childhood Lead) using various backbone architectures.
}
\label{fig:various_architectures}
\vspace{-.15in}
\end{figure}

%% file: figures/efficiency_sensitivity.tex
\begin{figure}[!t]
\centering
\includegraphics[width=\linewidth]{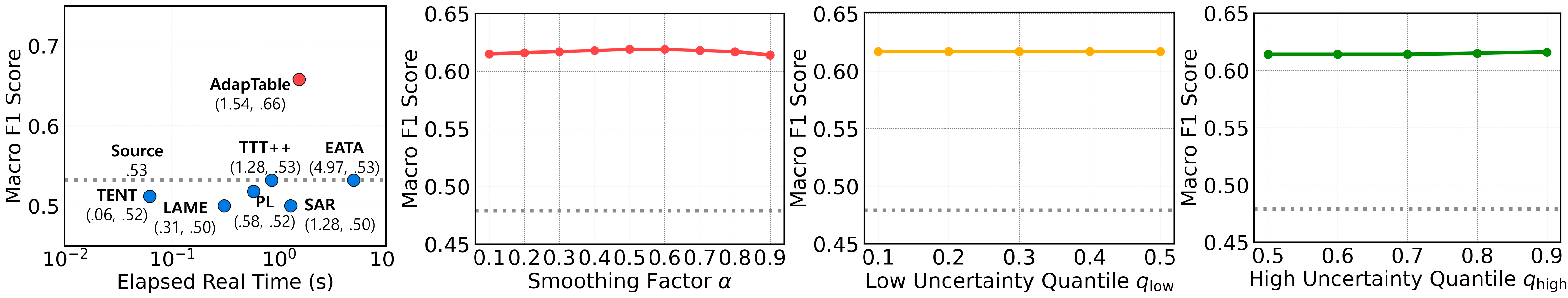}
\caption{
    Computational efficiency (leftmost figure) and hyperparameter sensitivity analysis of AdapTable (three figures on the right) using MLP on the HELOC and Childhood Lead datasets, respectively.
}
\label{fig:efficieny_sensitivity}
\vspace{-.2in}
\end{figure}

%% file: tables/further_common_corruption.tex
\begin{table}[!ht]
\centering
\caption{
    The average macro F1 score (\%) with their standard errors for TTA baselines is reported across six common corruptions---Gaussian, Uniform, Random Drop, Column Drop, Numerical, and Categorical---over three datasets---HELOC, Voting, and Childhood Lead. The results are averaged over three random repetitions.
}
\label{table:further_common_corruptions}
\begin{adjustbox}{width=\linewidth}
\begin{tabular}{clcccccc}
    \toprule

    Dataset & Method & Gaussian & Uniform & Random Drop & Column Drop & Numerical & Categorical \\ \midrule
    
    \multirow{7}{*}{HELOC} & Source & 33.1 $\pm$ 0.0 & 33.0 $\pm$ 0.0 & 31.4 $\pm$ 0.1 & 32.3 $\pm$ 1.4 & 33.8 $\pm$ 0.2 & 32.3 $\pm$ 0.3 \\

    & PL & 31.2 $\pm$ 0.0 & 31.2 $\pm$ 0.0 & 30.6 $\pm$ 0.0 & 31.1 $\pm$ 0.7 & 32.1 $\pm$ 0.2 & 30.4 $\pm$ 0.2 \\
    
    & TENT & 33.1 $\pm$ 0.0 & 33.0 $\pm$ 0.0 & 31.4 $\pm$ 0.1 & 32.3 $\pm$ 1.4 & 33.8 $\pm$ 0.2 & 32.3 $\pm$ 0.3 \\
    
    & EATA & 33.1 $\pm$ 0.0 & 33.0 $\pm$ 0.0 & 31.4 $\pm$ 0.1 & 32.3 $\pm$ 1.4 & 33.8 $\pm$ 0.2 & 32.3 $\pm$ 0.3 \\
    
    & SAR & 31.9 $\pm$ 0.1 & 32.0 $\pm$ 0.1 & 30.7 $\pm$ 0.2 & 31.3 $\pm$ 0.8 & 32.4 $\pm$ 0.4 & 31.4 $\pm$ 0.3 \\
    
    & LAME & 30.1 $\pm$ 0.0 & 30.1 $\pm$ 0.0 & 30.1 $\pm$ 0.0 & 30.1 $\pm$ 0.0 & 30.9 $\pm$ 0.1 & 29.4 $\pm$ 0.2 \\

    & AdapTable \cellcolor{brightgray} & \cellcolor{brightgray} \textbf{57.6 $\pm$ 0.1} \cellcolor{brightgray} & \cellcolor{brightgray} \textbf{57.8 $\pm$ 0.0} \cellcolor{brightgray} & \cellcolor{brightgray} \textbf{53.0 $\pm$ 0.1} \cellcolor{brightgray} & \cellcolor{brightgray} \textbf{52.1 $\pm$ 3.2} \cellcolor{brightgray} & \cellcolor{brightgray} \textbf{58.1 $\pm$ 0.1} \cellcolor{brightgray} & \cellcolor{brightgray} \textbf{58.9 $\pm$ 0.4} \cellcolor{brightgray} \\ \midrule

    \multirow{7}{*}{Voting} & Source & 76.6 $\pm$ 0.0 & 76.5 $\pm$ 0.0 & 72.5 $\pm$ 0.2 & 72.8 $\pm$ 0.4 & 76.3 $\pm$ 0.1 & \textbf{85.2 $\pm$ 0.1} \\

    & PL & 75.6 $\pm$ 0.3 & 75.2 $\pm$ 0.3 & 71.1 $\pm$ 0.5 & 70.6 $\pm$ 0.5 & 75.9 $\pm$ 0.1 & 85.1 $\pm$ 0.1 \\

    & TENT & 76.6 $\pm$ 0.0 & 76.5 $\pm$ 0.0 & 72.5 $\pm$ 0.2 & 72.8 $\pm$ 0.4 & 76.3 $\pm$ 0.1 & \textbf{85.2 $\pm$ 0.1} \\

    & EATA & 76.6 $\pm$ 0.0 & 76.5 $\pm$ 0.0 & 72.5 $\pm$ 0.2 & 72.8 $\pm$ 0.4 & 76.3 $\pm$ 0.1 & \textbf{85.2 $\pm$ 0.1} \\
    
    & SAR & 67.2 $\pm$ 1.0 & 64.0 $\pm$ 0.2 & 61.8 $\pm$ 1.0 & 60.8 $\pm$ 0.8 & 69.9 $\pm$ 0.1 & 84.2 $\pm$ 0.1 \\
    
    & LAME & 39.4 $\pm$ 0.2 & 39.4 $\pm$ 0.1 & 37.3 $\pm$ 0.0 & 37.8 $\pm$ 0.2 & 39.4 $\pm$ 0.2 & 81.4 $\pm$ 0.2 \\

    & AdapTable \cellcolor{brightgray} & \cellcolor{brightgray} \textbf{78.9 $\pm$ 0.0} \cellcolor{brightgray} & \cellcolor{brightgray} \textbf{78.6 $\pm$ 0.1} \cellcolor{brightgray} & \cellcolor{brightgray} \textbf{74.9 $\pm$ 0.1} \cellcolor{brightgray} & \cellcolor{brightgray} \textbf{75.5 $\pm$ 0.5} \cellcolor{brightgray} & \cellcolor{brightgray} \textbf{78.0 $\pm$ 0.1} \cellcolor{brightgray} & \cellcolor{brightgray} 85.0 $\pm$ 0.4 \cellcolor{brightgray} \\ \midrule

    \multirow{7}{*}{Childhood Lead} & Source & 47.9 $\pm$ 0.0 & 47.9 $\pm$ 0.0 & 47.9 $\pm$ 0.0 & 47.9 $\pm$ 0.0 & 48.1 $\pm$ 0.0 & 48.8 $\pm$ 0.0 \\
    
    & PL & 47.9 $\pm$ 0.0 & 47.9 $\pm$ 0.0 & 47.9 $\pm$ 0.0 & 47.9 $\pm$ 0.0 & 48.1 $\pm$ 0.0 & 48.8 $\pm$ 0.0 \\
    
    & TENT & 47.9 $\pm$ 0.0 & 47.9 $\pm$ 0.0 & 47.9 $\pm$ 0.0 & 47.9 $\pm$ 0.0 & 48.1 $\pm$ 0.0 & 48.8 $\pm$ 0.0 \\
    
    & EATA & 47.9 $\pm$ 0.0 & 47.9 $\pm$ 0.0 & 47.9 $\pm$ 0.0 & 47.9 $\pm$ 0.0 & 48.1 $\pm$ 0.0 & 48.8 $\pm$ 0.0 \\
    
    & SAR & 47.9 $\pm$ 0.0 & 47.9 $\pm$ 0.0 & 47.9 $\pm$ 0.0 & 47.9 $\pm$ 0.0 & 48.1 $\pm$ 0.0 & 48.8 $\pm$ 0.0 \\
    
    & LAME & 47.9 $\pm$ 0.0 & 47.9 $\pm$ 0.0 & 47.9 $\pm$ 0.0 & 47.9 $\pm$ 0.0 & 48.1 $\pm$ 0.0 & 48.8 $\pm$ 0.0 \\
    
    & AdapTable \cellcolor{brightgray} & \cellcolor{brightgray} \textbf{61.4 $\pm$ 0.1} \cellcolor{brightgray} & \cellcolor{brightgray} \textbf{61.5 $\pm$ 0.0} \cellcolor{brightgray} & \cellcolor{brightgray} \textbf{58.0 $\pm$ 0.1} \cellcolor{brightgray} & \cellcolor{brightgray} \textbf{55.9 $\pm$ 1.6} \cellcolor{brightgray} & \cellcolor{brightgray} \textbf{62.8 $\pm$ 0.2} \cellcolor{brightgray} & \cellcolor{brightgray} \textbf{53.1 $\pm$ 0.2} \cellcolor{brightgray} \\ 
    
    \bottomrule
\end{tabular}
\end{adjustbox}
\end{table}

%% file: tables/further_architecture.tex
\begin{table}[!t]
\centering
\caption{
    The average macro F1 score (\%) with their standard errors for TTA baselines is reported across three datasets---HELOC, Voting, and Childhood Lead---using three model architectures---AutoInt, ResNet, and FT-Transformer. The results are averaged over three random repetitions.
}
\label{table:further_architecture}
\begin{adjustbox}{width=.65\linewidth}
\begin{tabular}{clccc}
    \toprule

    Model & Method & HELOC & Voting & Childhood Lead \\ \midrule
    
    \multirow{7}{*}{AutoInt} & Source & 34.9 $\pm$ 0.0 & 77.5 $\pm$ 0.0 & 47.9 $\pm$ 0.0 \\
    
    & PL & 31.6 $\pm$ 0.0 & 76.5 $\pm$ 0.1 & 47.9 $\pm$ 0.0 \\
    
    & TENT & 34.9 $\pm$ 0.0 & 77.5 $\pm$ 0.0 & 47.9 $\pm$ 0.0 \\
    
    & EATA & 34.9 $\pm$ 0.0 & 77.5 $\pm$ 0.0 & 47.9 $\pm$ 0.0 \\
    
    & SAR & \textbf{62.0 $\pm$ 0.4} & 31.2 $\pm$ 0.7 & 47.9 $\pm$ 0.0 \\
    
    & LAME & 30.1 $\pm$ 0.0 & 37.3 $\pm$ 0.0 & 47.9 $\pm$ 0.0 \\
    
    & AdapTable \cellcolor{brightgray} & \cellcolor{brightgray} 56.3 $\pm$ 0.1 \cellcolor{brightgray} & \cellcolor{brightgray} \textbf{79.2 $\pm$ 0.0} \cellcolor{brightgray} & \cellcolor{brightgray} \textbf{61.8 $\pm$ 0.1} \cellcolor{brightgray} \\ \midrule

    \multirow{7}{*}{ResNet} & Source & 52.0 $\pm$ 0.0 & 76.6 $\pm$ 0.0 & 47.9 $\pm$ 0.0 \\
    
    & PL & 34.3 $\pm$ 0.1 & 73.3 $\pm$ 0.1 & 47.9 $\pm$ 0.0 \\
    
    & TENT & 52.0 $\pm$ 0.0 & 76.6 $\pm$ 0.0 & 47.9 $\pm$ 0.0 \\
    
    & EATA & 52.0 $\pm$ 0.0 & 76.7 $\pm$ 0.0 & 47.9 $\pm$ 0.0 \\
    
    & SAR & 55.1 $\pm$ 0.5 & 52.2 $\pm$ 0.5 & 47.9 $\pm$ 0.0 \\
    
    & LAME & 30.1 $\pm$ 0.0 & 75.1 $\pm$ 0.1 & 47.9 $\pm$ 0.0 \\
    
    & AdapTable \cellcolor{brightgray} & \cellcolor{brightgray} \textbf{61.9 $\pm$ 0.0} \cellcolor{brightgray} & \cellcolor{brightgray} \textbf{78.7 $\pm$ 0.0} \cellcolor{brightgray} & \cellcolor{brightgray} \textbf{61.3 $\pm$ 0.1} \cellcolor{brightgray} \\ \midrule
    
    \multirow{7}{*}{FT-Transformer} & Source & 33.0 $\pm$ 0.0 & 77.3 $\pm$ 0.0 & 47.9 $\pm$ 0.0 \\
    
    & PL & 30.6 $\pm$ 0.0 & 76.0 $\pm$ 0.1 & 47.9 $\pm$ 0.0 \\
    
    & TENT & 33.0 $\pm$ 0.0 & 77.3 $\pm$ 0.0 & 47.9 $\pm$ 0.0 \\
    
    & EATA & 33.0 $\pm$ 0.0 & 77.3 $\pm$ 0.0 & 47.9 $\pm$ 0.0 \\
    
    & SAR & 35.3 $\pm$ 0.1 & 73.6 $\pm$ 0.3 & 47.9 $\pm$ 0.0 \\
    
    & LAME & 30.7 $\pm$ 0.1 & 71.5 $\pm$ 0.1 & 47.9 $\pm$ 0.0 \\
    
    & AdapTable \cellcolor{brightgray} & \cellcolor{brightgray} \textbf{55.0 $\pm$ 0.0} \cellcolor{brightgray} & \cellcolor{brightgray} \textbf{79.2 $\pm$ 0.1} \cellcolor{brightgray} & \cellcolor{brightgray} \textbf{61.7 $\pm$ 0.1} \cellcolor{brightgray} \\
    
    \bottomrule
\end{tabular}
\end{adjustbox}
\end{table}

%% file: figures/further_latent_visualization.tex
\begin{figure*}[!ht]
\centering
\includegraphics[width=0.84\textwidth]{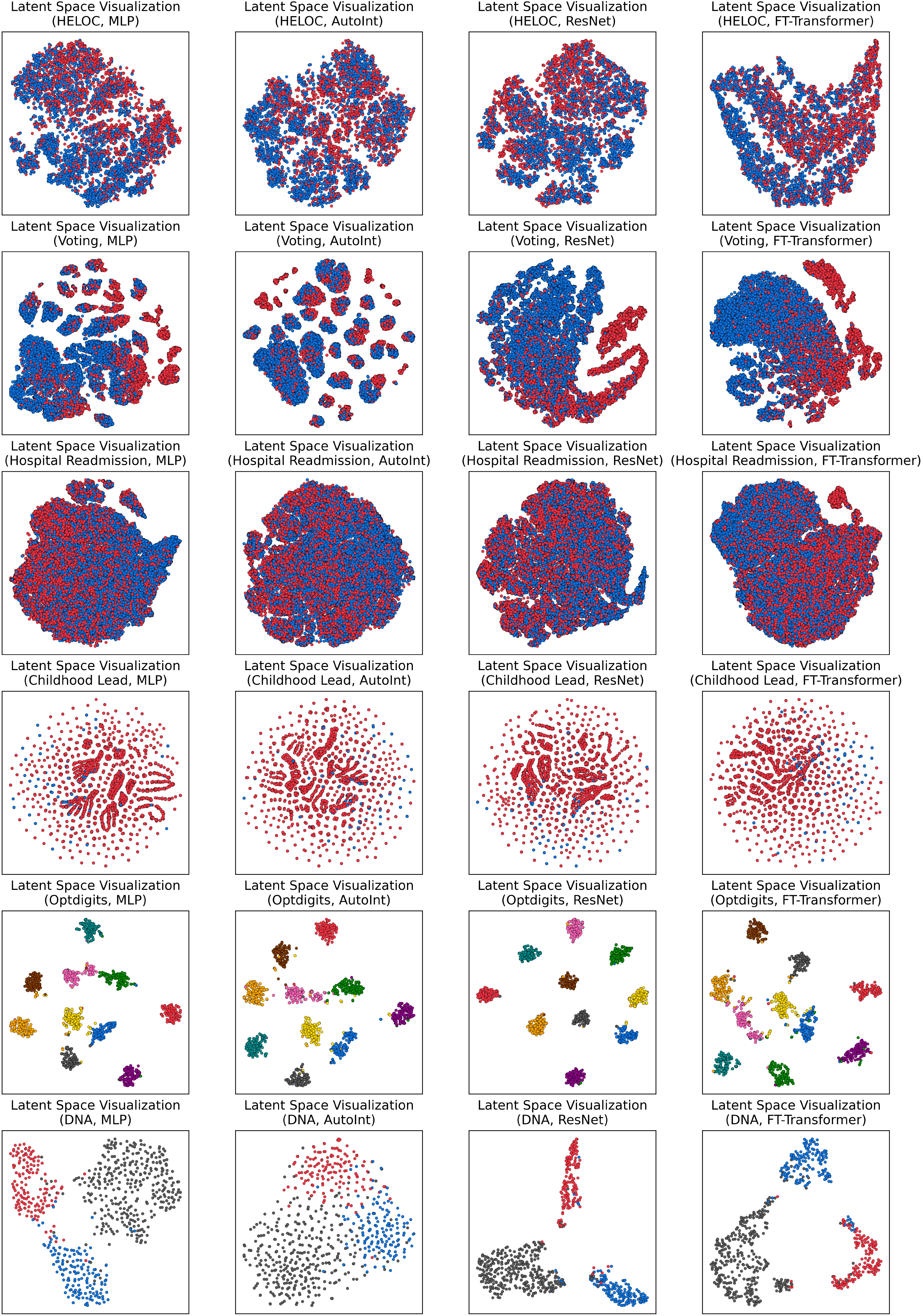}
\caption{
    Latent space visualizations of test samples using t-SNE across six diverse datasets, including tabular datasets (HELOC, Voting, Hospital Readmission, and Childhood Lead) and non-tabular datasets (Optdigits, DNA), applied to various deep tabular learning architectures.
}
\label{fig:further_latent}
\end{figure*}

%% file: figures/further_reliability_diagram.tex
\begin{figure*}[!ht]
\centering
\includegraphics[width=\textwidth]{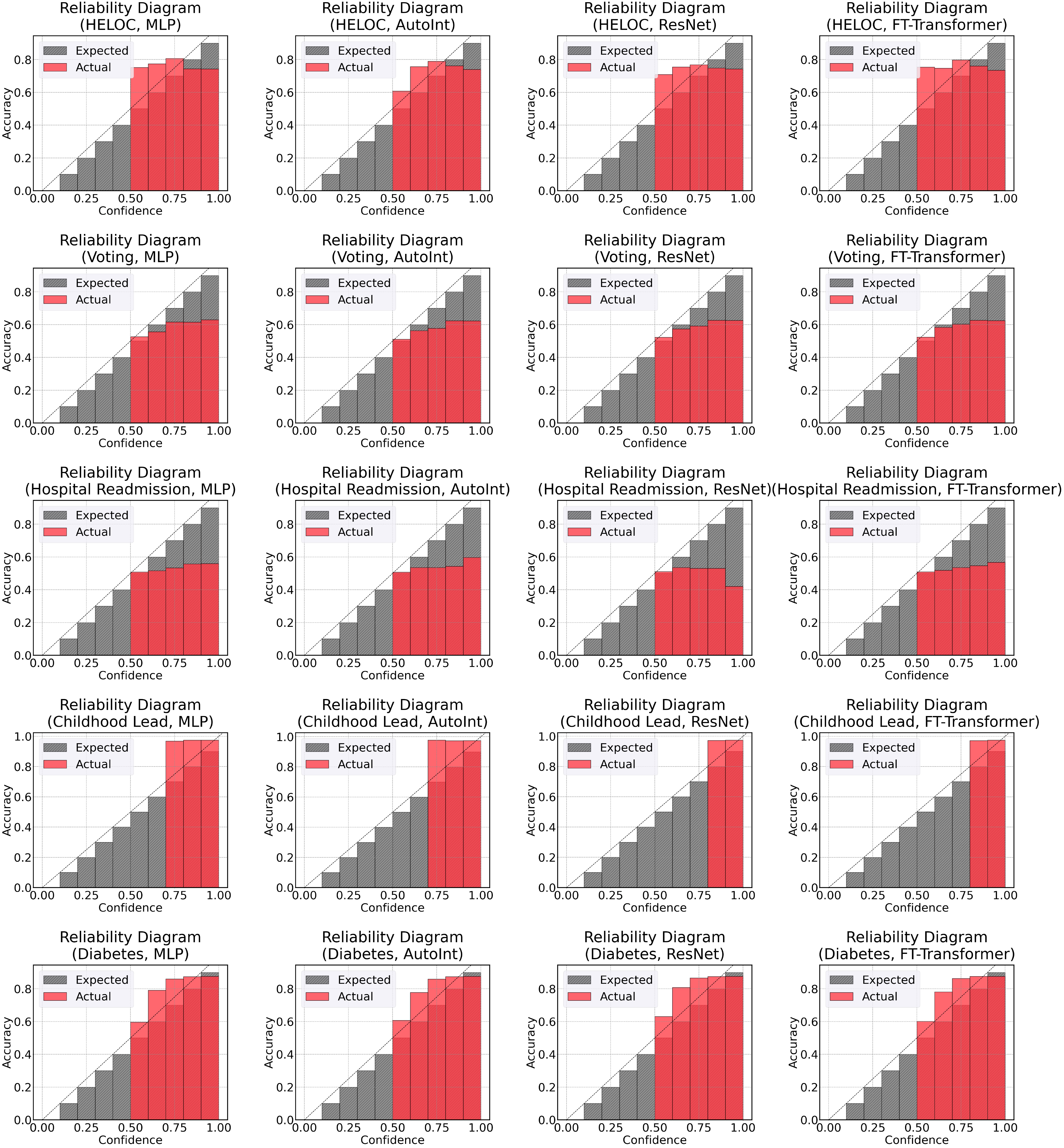}
\caption{
    Reliability diagrams for test instances across five different tabular datasets (HELOC, Voting, Hospital Readmission, Childhood Lead, and Diabetes) and four representative deep tabular learning architectures (MLP, AutoInt, ResNet, FT-Transformer).
}
\label{fig:further_reliability}
\end{figure*}

%% file: figures/further_label_distribution.tex
\begin{figure*}[!ht]
\centering
\includegraphics[width=\textwidth]{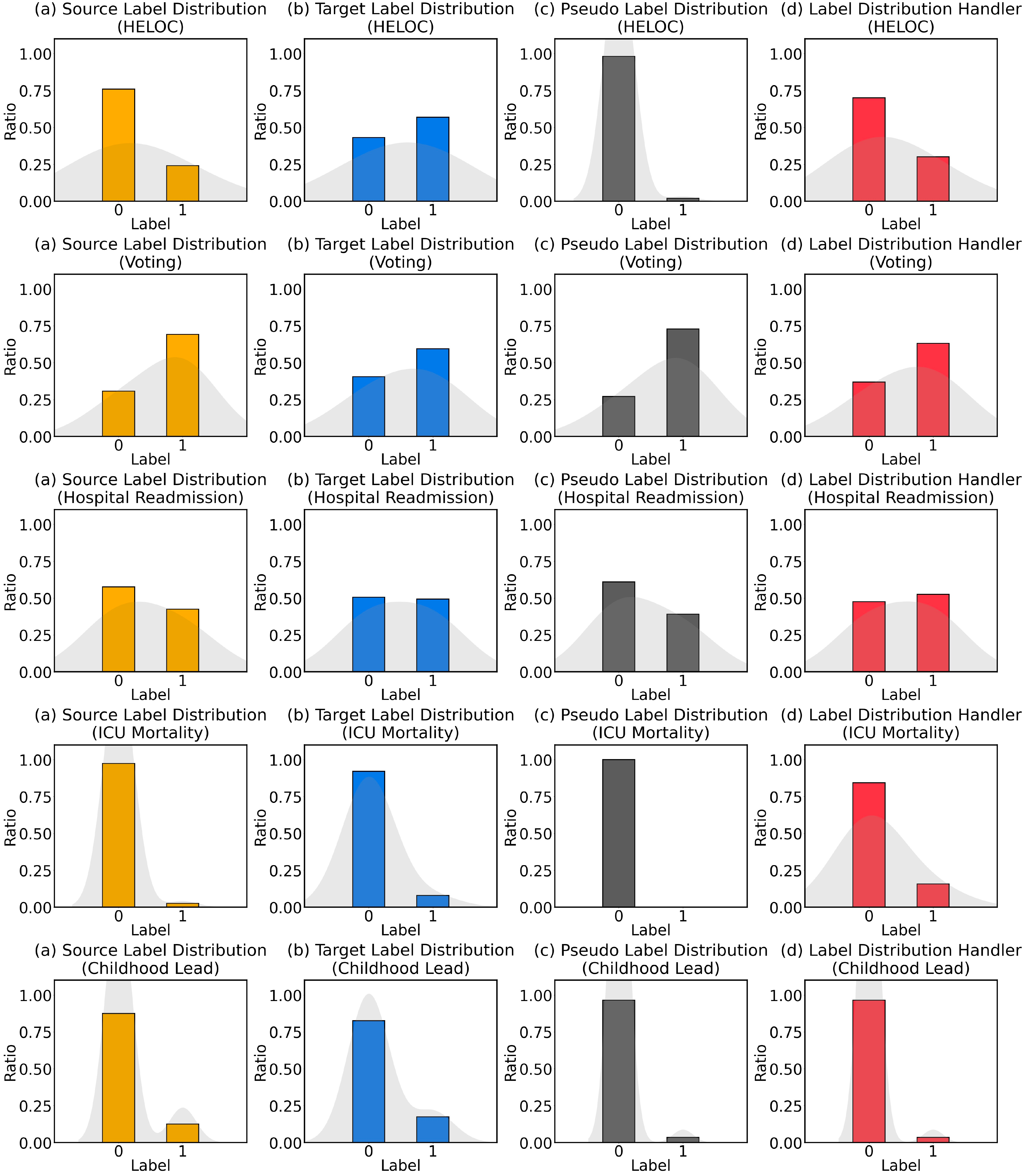}
\caption{
    Label distribution histograms for test instances showing (a) source label distribution, (b) target label distribution, (c) pseudo label distribution, and (d) estimated target label distribution after applying our label distribution handler, across five tabular datasets (HELOC, Voting, Hospital Readmission, Childhood Lead, and Diabetes) using MLP.
}
\label{fig:further_label_distribution}
\end{figure*}

%% file: figures/further_entropy_distribution.tex
\begin{figure*}[!ht]
\centering
\includegraphics[width=0.87\textwidth]{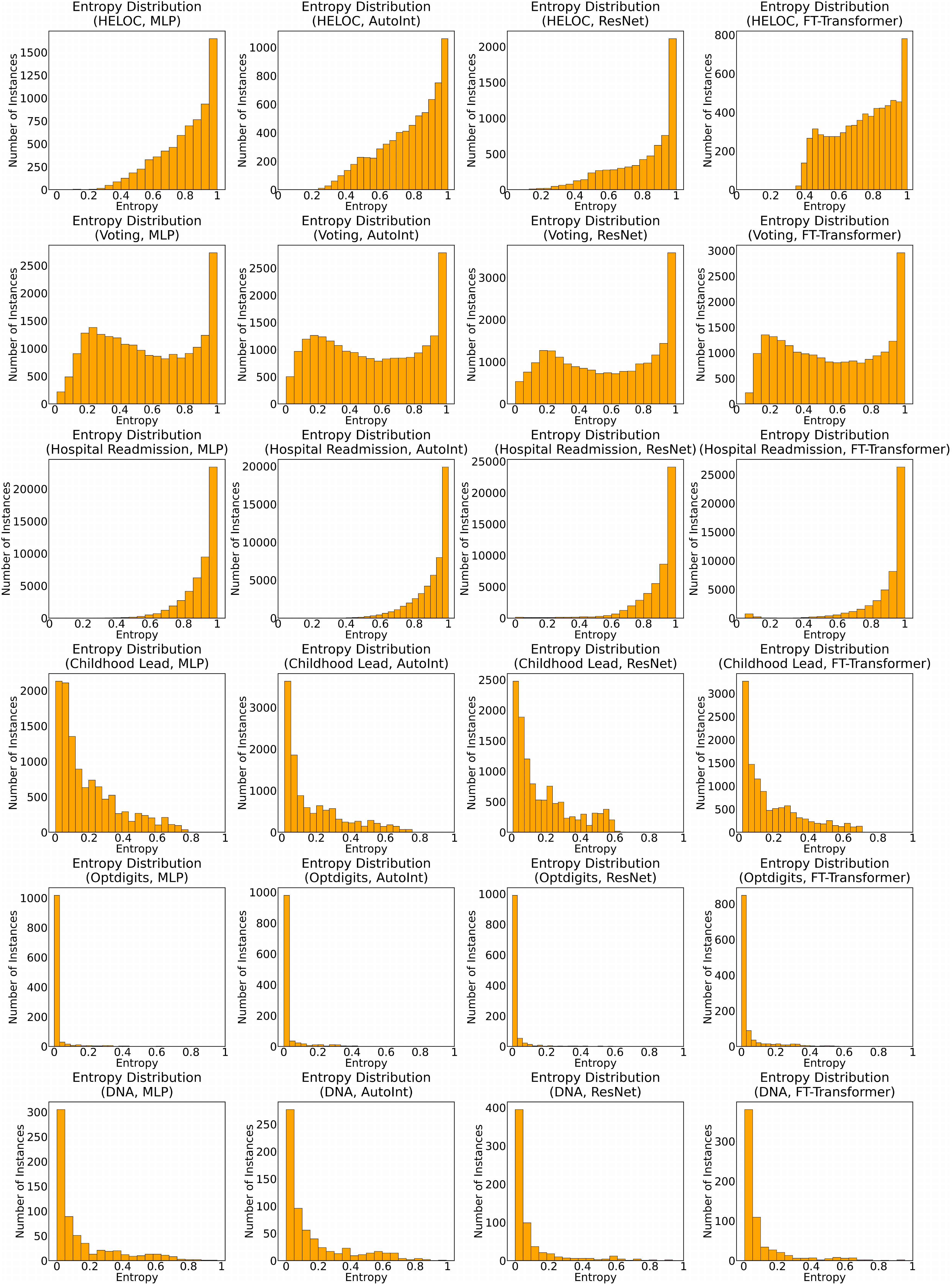}
\caption{
    Entropy distribution histograms of test samples across six diverse datasets, including tabular datasets (HELOC, Voting, Hospital Readmission, and Childhood Lead) and non-tabular datasets (Optdigits, DNA), applied to four deep tabular learning architectures. Prediction entropies are normalized by dividing by the maximum entropy, $\log{C}$, where $C$ represents the number of classes for each dataset.
}
\label{fig:further_entropy}
\end{figure*}